\documentclass[]{alaya}
\usepackage{makecell}
\usepackage{wrapfig}
\usepackage{tabularx}
\usepackage{textcomp}
\usepackage{stfloats}
\usepackage{url}
\usepackage{verbatim}
\usepackage{titlesec}
\usepackage{adjustbox}
\usepackage{multirow}
\usepackage{pifont}
\usepackage[sc]{mathpazo}
\usepackage{tikz}
\usepackage{amsmath,amssymb}
\usepackage{colortbl}
\usepackage[numbers,sort&compress]{natbib}
\usepackage{booktabs}
\usepackage{hyperref}
\usepackage{graphicx}
\RequirePackage{xspace}
\makeatletter
\DeclareRobustCommand\onedot{\futurelet\@let@token\@onedot}
\def\@onedot{\ifx\@let@token.\else.\null\fi\xspace}
\usepackage[most]{tcolorbox}
\usepackage{array}
\usepackage{siunitx}
\usepackage{caption}
\definecolor{headerpurple}{HTML}{d8d2fc}
\definecolor{rowgray}{gray}{0.95}
\usepackage{CJKutf8}
\def\eg{\emph{e.g}\onedot}

\makeatother

\definecolor{adptorange}{RGB}{248, 205, 172}
\definecolor{cmpblue}{RGB}{189, 215, 238}

\definecolor{our_red}{RGB}{232,157,160}
\definecolor{our_blue}{RGB}{136,206,230}
\definecolor{our_orange}{RGB}{246,200,168}
\definecolor{our_green}{RGB}{178,211,164}

\definecolor{mygray}{HTML}{f0f0f0}

\usepackage{bbding}
\usepackage{fontawesome}
\usepackage{float}
\usepackage{flafter}
\usepackage{algorithm}
\usepackage{algorithmic}
\usepackage{listings}
\definecolor{lstbg}{gray}{0.96}
\definecolor{lstcomment}{rgb}{0.35,0.45,0.40}
\newcommand{\cmark}{\textcolor{green!70!black}{\ding{51}}}

\newlength\savewidth

\newcolumntype{x}[1]{>{\centering\arraybackslash}p{#1pt}}
\newcolumntype{y}[1]{>{\raggedright\arraybackslash}p{#1pt}}
\newcolumntype{z}[1]{>{\raggedleft\arraybackslash}p{#1pt}}

\renewcommand{\paragraph}[1]{\vspace{1.25mm}\noindent\textbf{#1}}

\definecolor{green}{HTML}{009000}
\definecolor{red}{HTML}{ea4335}

\newcommand{\methodname}{MASS}

\newcommand{\ntick}{\ensuremath{N}}
\newcommand{\cclients}{\ensuremath{C}}

\newcommand{\sharedstate}[1]{\ensuremath{\hat{s}_{#1}}}
\newcommand{\statespace}{\ensuremath{\mathcal{S}}}
\newcommand{\jointaction}[1]{\ensuremath{a_{#1}}}
\newcommand{\transitionmodel}{\ensuremath{F_\theta}}
\newcommand{\renderermodel}{\ensuremath{R_\phi}}

\title{\methodname{}: Multiplayer World Models with Authoritative Shared State}
\renewcommand\authorformat[2][]{\mbox{\sffamily\bfseries #2$^{#1}$}}
\renewcommand{\authorlist}{%
  \makebox[\linewidth][l]{%
    \authorformat[1,2,3,\S]{Ziqi Cai},
    \authorformat[1,2,\S]{Siqi Yang},
    \authorformat[2]{Yimu Wang},
    \authorformat[1,\S]{Zixian Gao},
    \authorformat[2]{Yunheng Liu},%
  }\par
  \makebox[\linewidth][l]{%
    \authorformat[2]{Shuchen Weng},
    \authorformat[3]{Erwin Wu},
    \authorformat[1,\ddagger]{Kaipeng Zhang},
    \authorformat[2,\dagger]{Boxin Shi}%
  }%
}
\affiliation[1]{Alaya Lab}
\affiliation[2]{Peking University}
\affiliation[3]{Institute of Science Tokyo}

\abstract{
Current video world models struggle in multiplayer environments because they entangle world state with view-dependent visual latents, leading to redundant compute, view inconsistencies, and poor scalability. We propose \methodname{} (\textbf{M}ultiplayer world models with \textbf{A}uthoritative \textbf{S}hared \textbf{S}tate) to resolve this limitation. Inspired by multiplayer game architectures, \methodname{} disentangles world dynamics and view rendering. A learned Logic Engine advances a global, authoritative typed state from joint actions without any hand-written transition function, acting as the sole recurrent memory and synchronization reference. From this shared state, a learned Rendering Engine generates independent and consistent views for any requested camera on demand. This explicit disentangling allows \methodname{} to achieve superior state accuracy and lower cross-view inconsistency compared to state-of-the-art multi-view baselines on a matched multiplayer Snake benchmark. It advances predicted worlds with 1,024 concurrent players for 10,000 recurrent steps. Our results show that explicit, authoritative state modeling provides a practical foundation for scalable and consistent multi-agent world simulation.
}

\github{\url{https://alaya-lab.github.io/MASS/}}
\date{\today}

\begin{document}
\maketitle
\begingroup
\renewcommand{\thefootnote}{\fnsymbol{footnote}}
\footnotetext[4]{This work was done during their internships at Alaya Lab.}
\footnotetext[2]{Corresponding author. \email{shiboxin@pku.edu.cn}.}
\footnotetext[3]{Project leader. \email{kaipeng.zhang@shanda.com}.}
\endgroup

\begingroup
    \centering
    \includegraphics[width=\linewidth]{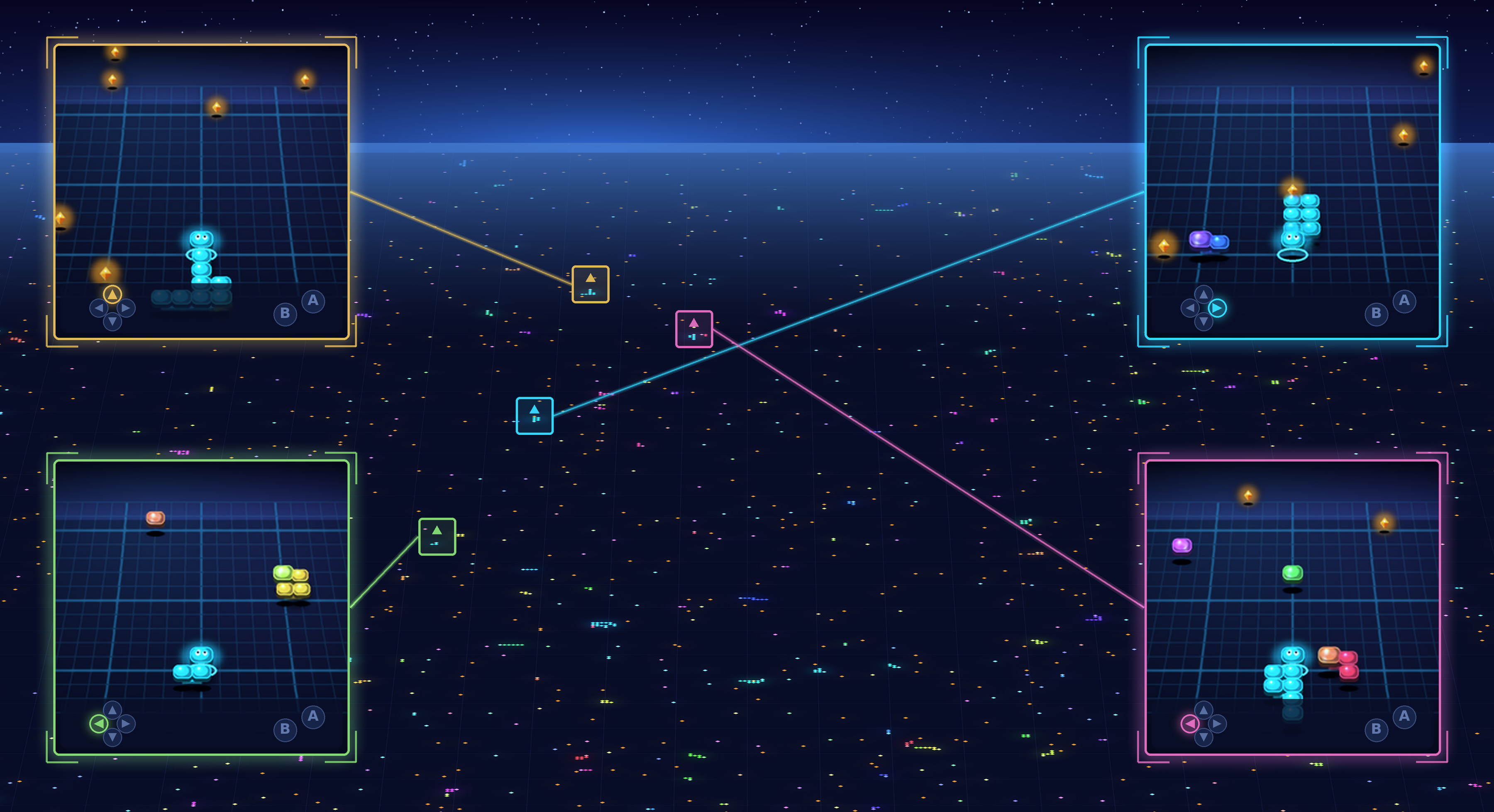}\vspace{5pt}
    \captionsetup{hypcap=false}
    \captionof{figure}{A predicted Snake world with 1{,}024 players.
    The center visualizes the authoritative typed state. Four screens show learned
    renders for selected client cameras and the gamepads show their current actions.}
    \label{fig:teaser}
\endgroup

\section{Introduction}
\label{sec:intro}

Interactive video world models simulate video games by predicting the next
observation from world history and the player's action.
GameNGen~\citep{valevski2024gamengen}, Oasis~\citep{oasis2024},
DIAMOND~\citep{alonso2024diamond}, Genie~\citep{bruce2024genie}, and
WHAM~\citep{kanervisto2025wham} produce responsive, visually detailed rollouts
without running a traditional game engine. Their pixel or visual-latent interface combines
one player's view with the recurrent memory used for simulation, which suits a
single-player setting.

Multiplayer simulation needs a recurrent state that belongs to the world rather
than to any camera. A thousand players can inhabit \emph{one} world,
yet each camera reveals only a small part of it. Separate recurrent visual
histories encode the same shared content a thousand times, allow simultaneous
views to disagree about the same entity, and tie the cost of \emph{simulating}
the world to the number of views being \emph{watched}. Existing multiplayer
world models coordinate views through joint generation or shared visual
representations.
Gamma-World~\citep{gammaworld2026}, MultiWorld~\citep{multiworld2026},
MultiGen~\citep{multigen2026}, and WanToFight~\citep{wantofight2026} keep
their shared representations visual, dense, or externally maintained.
None use a typed authoritative state that serves as both recurrent memory
and synchronization object.

Online games already separate shared simulation from per-player presentation
\citep{bernier2001latency,claypool2006latency}. An authoritative server
advances one canonical game state. Clients receive versioned snapshots,
predict locally through update gaps, and render their own cameras. The world
is simulated \emph{once}, independent of the number of clients that render it.
\methodname{} (\textbf{M}ultiplayer world models with \textbf{A}uthoritative
\textbf{S}hared \textbf{S}tate) brings this contract to learned world models.

In \methodname{}, a \emph{schema} declares the typed entities of a game. At
each tick, a learned \textbf{Logic Engine} advances the complete typed state
from the joint player actions and declared exogenous inputs. Every game in this
report instantiates the same decoder-only Transformer architecture with
game-specific tokenizers, embeddings, and model weights. A learned
\textbf{Rendering Engine} then synthesizes each requested observation from
that predicted state and a client camera. The typed state carries the recurrent
memory and is published as the client synchronization message. Every rendered
view conditions on that same state. No hand-written transition function runs
during rollout. \Cref{fig:teaser} shows one predicted Snake world state with four client
views decoded from it.

Our contributions can be summarized as:
\begin{itemize}
\item We formulate learned multiplayer simulation as prediction of an
authoritative typed state followed by camera-conditioned rendering. The same
predicted state is the recurrent history, the synchronization reference, and
the source of every client view.
\item We develop a schema-based architecture where the Logic Engine
advances typed entity records and the Rendering Engine generates requested
observations. Rollout uses no native game transition, and the Logic Engine can
temporarily predict from a synchronized client state during update gaps.
\item We evaluate this formulation with matched multiplayer baselines,
cross-game studies, simulated update stalls, and long recurrent rollouts.
\methodname{} reaches 0.76 state recovery on matched Snake compared with 0.128
for the strongest video-based baseline, and it advances 1{,}024 simulated
player entities for 10{,}000 recurrent ticks.
\end{itemize}

\section{Related work}
\label{sec:related}

\paragraph{Video world models.}
GameNGen~\citep{valevski2024gamengen}, Oasis~\citep{oasis2024},
DIAMOND~\citep{alonso2024diamond}, Genie~\citep{bruce2024genie}, and
WHAM~\citep{kanervisto2025wham} learn interactive environments by recurrently
predicting action-conditioned observations, using pixels or visual latents as
the recurrent carrier
\citep{oh2015action,chiappa2017recurrent,kim2020gamegan,bamford2020neural,
menapace2021playable,yang2024unisim}. Each new frame is decoded from a
recurrent state that also serves as the input for the next step, which makes
the simulation responsive in single-player settings. In multiplayer settings,
however, every requested view would need its own independent recurrent
history, and those histories can drift apart for the same world. \methodname{}
avoids this by predicting one shared typed state first, then decoding any
number of views from that single prediction.

\paragraph{Multiplayer generation.}
Gamma-World~\citep{gammaworld2026}, MultiWorld~\citep{multiworld2026},
MultiGen~\citep{multigen2026}, Agora-1~\citep{agora2026}, and
WanToFight~\citep{wantofight2026} introduce global visual representations
(MultiWorld), sparse hub attention with linear cross-agent cost
(Gamma-World), or persistent external memories independent of the context
window (MultiGen). ShareVerse~\citep{zhu2026shareverse},
MetaWorld~\citep{hu2026metaworld}, and Prisma-World~\citep{sun2026prismaworld}
further couple multi-agent video streams through cross-agent attention,
world-state alignment, or geometry-aware joint denoising.
Solaris~\citep{solaris2026} targets consistent multiplayer
views in Minecraft using synchronized multi-agent videos and actions. It uses
staged single-player-to-multiplayer training and Checkpointed Self Forcing to
enable a longer-horizon teacher. Hu et al.~\citep{hu2026multiplayer} model
four-player Rocket League with a 5B latent diffusion model conditioned on
multiple action streams. WorldWeaver~\citep{mo2026worldweaver} augments
streaming video diffusion with latent cross-agent world-state registers, while
the Khora technical preview~\citep{khora2026} uses a shared STBoard for action
updates and synchronized rendering. \methodname{} differs by making the
recurrent world state explicitly typed, schema-constrained, directly evaluable
before rendering, and suitable as a versioned synchronization object. The
simulation cost stays independent of the number of rendered cameras because
every view decodes from the same predicted state.

\paragraph{Structured dynamics.}
Ha and Schmidhuber~\citep{ha2018world} showed that world models can be learned
in compact latent spaces without predicting every pixel directly. Recurrent
state-space predictors~\citep{hafner2019planet,hafner2020dreamer,
hafner2023dreamerv3}, neural cellular automata~\citep{mordvintsev2020nca}, and
convolutional recurrent models~\citep{shi2015convlstm} further demonstrated
that dynamics can be learned in state spaces beyond raw pixels
\citep{schrittwieser2020muzero,micheli2023iris,robine2023twm,
sanchez2020gns}. These approaches use dense or latent representations that
lack explicit entity structure, making it difficult to measure whether the
predicted state is semantically valid before rendering it. WRBench
\citep{wrbench2026} reports that recent video world models often fail to advance
events while they are outside the camera view, further motivating recurrent
state representations that are decoupled from observation. With the typed
records in \methodname{}, entity persistence, position accuracy, and structural
validity can be measured directly on the predicted state without generating a
single frame.

\paragraph{Multiplayer execution.}
Networked games separate an authoritative world update from state replication
and client presentation~\citep{bernier2001latency,claypool2006latency}. The
server advances one canonical game state, while each client renders its own
view from versioned snapshots. \methodname{} brings this same separation to
learned world models. The learned transition runs once per tick, and the
rendering stage serves any number of cameras from that single predicted state.
This decoupling keeps the simulation workload independent of the spectator
count, matching the efficiency model of traditional game servers.

\section{Method}
\label{sec:method}

\begin{figure*}[t]
\centering
\includegraphics[width=1\textwidth]{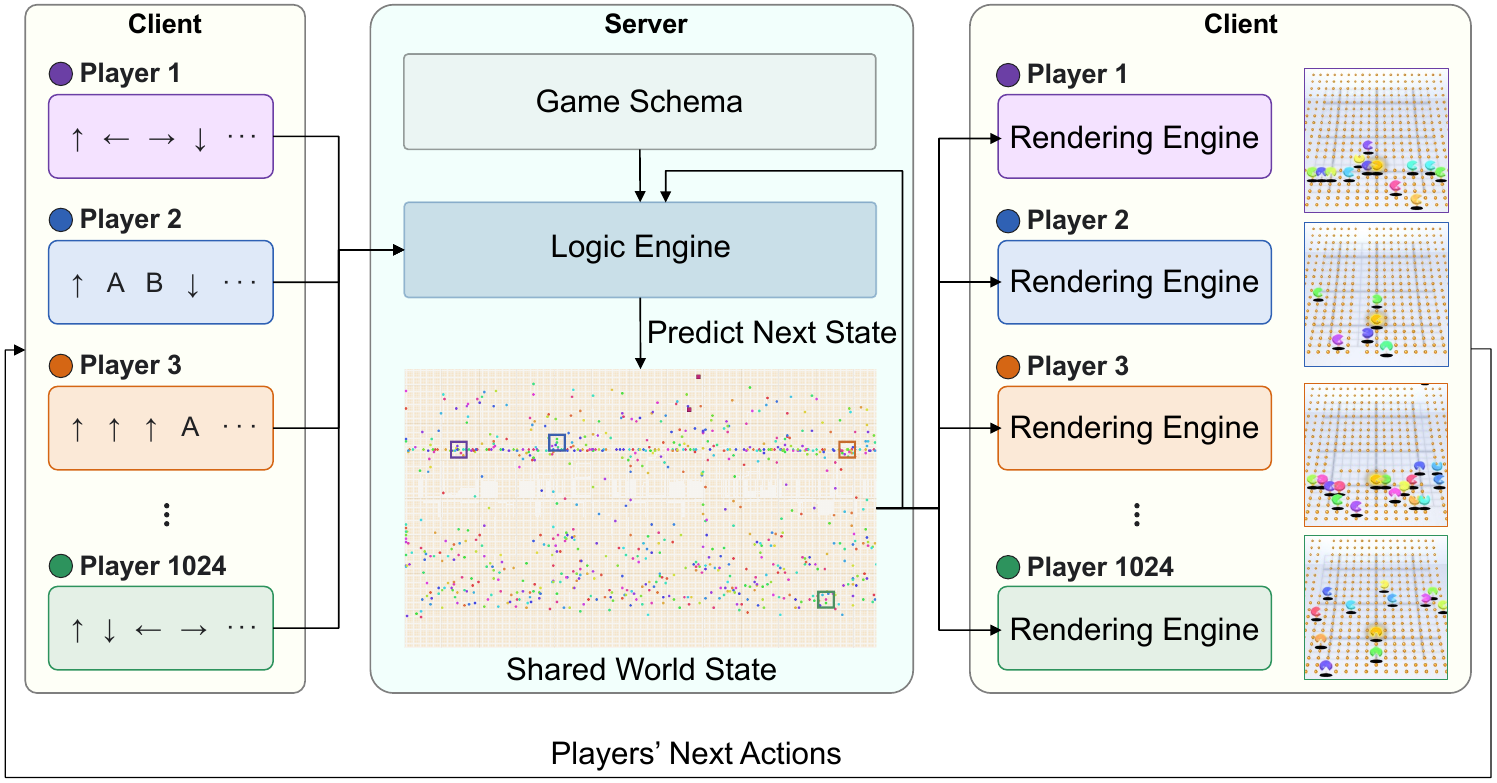}
\caption{Overview of \methodname{}. Clients send player actions to the server each tick. The game schema defines the typed state, and the learned Logic Engine advances it from the joint actions to produce the next world state. The updated state is distributed to clients, where each Rendering Engine generates a player view from that shared state and the corresponding camera. Players then send their next actions, and the loop repeats. The world transition runs once per tick, while views are rendered on demand.}
\label{fig:method}
\end{figure*}

\methodname{} uses the predicted typed state of the video game world as its single
recurrent object. First, a World State Tokenizer maps the state to records declared
by a game schema
(\Cref{sec:typed-shared-state}). Then, the Logic Engine advances the records from
the current state, the joint player actions, and declared exogenous inputs
(\Cref{sec:logic-engine}). Next, the Rendering Engine converts camera-local
projections of the predicted state into frames (\Cref{sec:state-renderer}).
During rollout, both the transitions and the
observations are produced by neural networks. The predicted state is
versioned and distributed to clients. During an update stall (\eg, a network interruption), the same Logic
Engine can temporarily advance a client's latest version locally to maintain responsiveness
(\Cref{sec:authoritative-distribution}). We show an overview in \Cref{fig:method}.

\subsection{World State Tokenizer}
\label{sec:typed-shared-state}

Most world models carry recurrent information in a video latent, which mixes
world dynamics with appearance. \methodname{} instead uses an explicit typed
state as the recurrent variable. Named fields make entities and their
relations directly inspectable. A canonical serialization also provides a
concrete, image-free synchronization message. Each new game supplies a schema
and trajectories. The core architecture stays the same, but each game
trains its own tokenizer, embeddings, and model weights.
Our implementation uses bounded, schema-defined layouts so that record
sequences can be batched efficiently and decoded with position-specific field
masks. Canonical padding and type-tag ranges map every valid state to a
fixed-width sequence, so the model can process any state admitted by the same
schema without per-game architecture changes.

The schema of a game is a short list of declarations, one per kind of entity,
\begin{equation}
\Sigma_g=\bigl\{(\text{kind}_j,\ \text{fields}_j,\ \text{count}_j)\bigr\}_{j=1}^{J_g}.
\label{eq:schema}
\end{equation}
Each entry names one kind of entity, the fields that describe it with the
values each field may take, and the number of instances the world holds. The
schema defines representation and validity, while the learned model supplies
the transition.

At every tick, the world state contains one record for each entity declared by
the schema. Each record holds the current values of that entity's fields,
\begin{equation}
s_t=\bigl(\rho_t^{1},\ldots,\rho_t^{K_g}\bigr),
\qquad K_g=\textstyle\sum_{j}\text{count}_j.
\label{eq:typed-state}
\end{equation}
We write $\statespace_g$ for the set of states the schema allows and
$\mathcal{A}_{\Sigma_g}$ for the assembly step that reads a full set of
records back into one state.

The World State Tokenizer turns each record into one short token sequence,
\begin{equation}
u_t^i=T_{\Sigma_g}\!\left(\rho_t^i,\ c_t^i,\ a_t^{\pi(i)},\ e_t^i\right),
\label{eq:record-tokenizer}
\end{equation}
where $c_t^i$ is a small window of nearby cells built from the current state
and joint actions, $a_t^{\pi(i)}$ is the action of the player attached to the
entity when one exists, and $e_t^i$ is the share of recorded randomness that
belongs to this record. Numbers, coordinates, flags, optional values, and
lists are written with type tags at fixed positions, so every record decodes
back into its fields without loss.

In Snake, for example, the schema has three entries. The global entry holds
the tick counter. Each snake entry records the body segments, heading, and life
status of one of the 1{,}024 snakes. Each of the 4{,}096 food entries lists
the food cells inside one $8\times8$ region. One tick therefore instantiates
$K_g=5{,}121$ records.

\subsection{Logic Engine}
\label{sec:logic-engine}

The Logic Engine is the only learned component that advances the shared state.
It applies one decoder-only Transformer to the tokenized records.
The model interface is shared across all record types, and each game trains its own tokenizer, embeddings, and model weights with the same core architecture. Let
$\sharedstate{t}\in\statespace_g$ be the predicted world state at
tick $t$, $\jointaction{t}$ the joint action of all players, and $e_t$ the
exogenous input. The world-level transition is
\begin{equation}
\sharedstate{t+1}=\transitionmodel(\sharedstate{t},\jointaction{t},e_t).
\label{eq:transition}
\end{equation}
The tokenizer and assembler realize it through record-wise prediction. If
$y_{t+1}^{i,j}$ is token $j$ of the next state of record $i$, the learned
distribution factorizes as
\begin{equation}
p_\theta(s_{t+1}\mid s_t,a_t,e_t)
\approx\prod_{i=1}^{K_g}\prod_j
p_\theta\!\left(y_{t+1}^{i,j}\mid u_t^i,y_{t+1}^{i,<j}\right),
\label{eq:typed-transition}
\end{equation}
where each record conditions on its own prefix and its previously decoded
tokens. Self-attention stays confined to each record's sequence. Interactions
between world objects enter through the neighborhood window $c_t^i$, computed
from the shared state before prediction. Records are batched freely at
training and inference with no attention between them. This factorization
supports the 1{,}024-entity worlds in our experiments.

A schema-derived mask restricts every output position to the values its
field allows, and a deterministic selector enforces cross-record constraints
(sortedness, uniqueness) during decoding. The assembler produces
\begin{equation}
\hat{s}_{t+1}=\mathcal{A}_{\Sigma_g}
\left(\{\hat{\rho}_{t+1}^{1},\ldots,\hat{\rho}_{t+1}^{K_g}\}\right),
\label{eq:state-assembly}
\end{equation}
which becomes the next recurrent input and the single state read by every
downstream view.

\begin{table}[t]
\centering
\small
\setlength{\tabcolsep}{4pt}
\caption{Architectural comparison of multiplayer world models. B-PV: per-view video predictor; B-SL: shared-latent predictor; B-UN: dense joint-state predictor.}
\label{tab:architecture-comparison}
\begin{adjustbox}{width=\textwidth}
\begin{tabular}{llllll}
\toprule
Method & Recurrent carrier & Explicit shared state & Logic and rendering
& Synchronization object & View execution \\
\midrule
B-PV & RGB for each view & N/A & N/A & N/A & Per view \\
B-SL & Visual latent & N/A & N/A & N/A & Joint latent \\
MultiWorld~\citep{multiworld2026} & Coupled multiview & N/A & N/A & N/A & Joint views \\
B-UN & Dense state grid & \cmark & Separate & Dense grid & On demand \\
\methodname{} & Typed state & \cmark & Separate & Full state & On demand \\
\bottomrule
\end{tabular}
\end{adjustbox}
\end{table}

\begin{table}[t]
\centering
\small
\caption{Matched multiplayer Snake results. Best in bold and second best
underlined.}
\label{tab:matched}
\begin{tabular*}{\columnwidth}{@{\extracolsep{\fill}}lccccc@{}}
\toprule
Metric & \methodname{} & MultiWorld & B-PV & B-SL & B-UN \\
\midrule
LPIPS $\downarrow$ & \textbf{0.098} & 0.277 & 0.397 & 0.396 & \underline{0.123} \\
Parser $\uparrow$ & \textbf{0.764} & 0.067 & \underline{0.128} & 0.083 & 0.000 \\
Count $\uparrow$ & \textbf{0.234} & 0.006 & \underline{0.141} & 0.109 & 0.051 \\
Pos. $\uparrow$ & \textbf{0.355} & 0.250 & \underline{0.286} & 0.180 & 0.002 \\
Event F1 $\uparrow$ & \textbf{0.552} & 0.065 & \underline{0.542} & 0.528 & 0.007 \\
X-view $\downarrow$ & \textbf{0.000} & 0.984 & 1.000 & 1.000 & \underline{0.051} \\
Invalid $\downarrow$ & \underline{0.177} & 0.981 & \textbf{0.052} & \textbf{0.052} & 1.000 \\
\bottomrule
\end{tabular*}
\end{table}

\subsection{Rendering Engine}
\label{sec:state-renderer}

Rendering begins from a world whose structure is already explicit. For
camera $i$, a projection $P_i$ writes what occupies each visible cell, which
entity it belongs to, which player owns the camera, and the view geometry
into a camera-local tensor. A learned residual U-Net turns this
projection into an RGB frame,
\begin{equation}
\hat{o}_t^i=\renderermodel(P_i(\sharedstate{t}),\kappa_i).
\label{eq:render}
\end{equation}
For the requested camera set $\mathcal{C}_t$, the observation model
factorizes as
\begin{equation}
p_\phi(\hat{o}_t^{\mathcal{C}_t}\mid\sharedstate{t},\kappa_{\mathcal{C}_t})
=\prod_{i\in\mathcal{C}_t}
p_\phi(\hat{o}_t^i\mid P_i(\sharedstate{t}),\kappa_i).
\label{eq:view-factorization}
\end{equation}
Every view reads the same shared state, so the decoder controls appearance
without reconciling separate visual histories. The renderer reads the
projected typed state and camera rather than an RGB history. Cameras can be
added, moved, or upgraded at rollout time without changing the world
transition. Resolution, materials, and lighting can likewise change without
retraining the learned dynamics. The renderers in this report share the typed
interface across different visual themes.
This separation of world dynamics from view synthesis can be evaluated independently through the State-Referenced Semantic Correctness (SRSC) protocol described in \S\ref{sec:eval}.

\subsection{Server and client logic}
\label{sec:authoritative-distribution}

\emph{Authoritative} denotes the sole state version adopted by all clients,
regardless of prediction accuracy. After each global transition the server
publishes a versioned typed snapshot. Clients that apply the same version
render from the same predicted world. With \ntick{} active entities,
\cclients{} state recipients, and snapshot size
$S$, the evaluated per-tick server workload is
\begin{equation}
T_{\mathrm{server}}=
T_{\mathrm{logic}}(\ntick)
+\cclients\, t_{\mathrm{sync}}(S).
\label{eq:cost}
\end{equation}
The learned transition runs \emph{once}. Distribution is linear in
\cclients{}, while rendering runs locally on each client from the received
typed state. The
versioned state also initializes temporary client-side prediction during a
stall. The client advances the latest authoritative state with the same Logic
Engine, supplies its own actions, and leaves other players' actions and unknown
spawns empty until the next server version arrives.

\section{Experiments}
\label{sec:eval}
\setcounter{topnumber}{2}

\begin{figure}[t]
\centering
\includegraphics[width=1\textwidth]{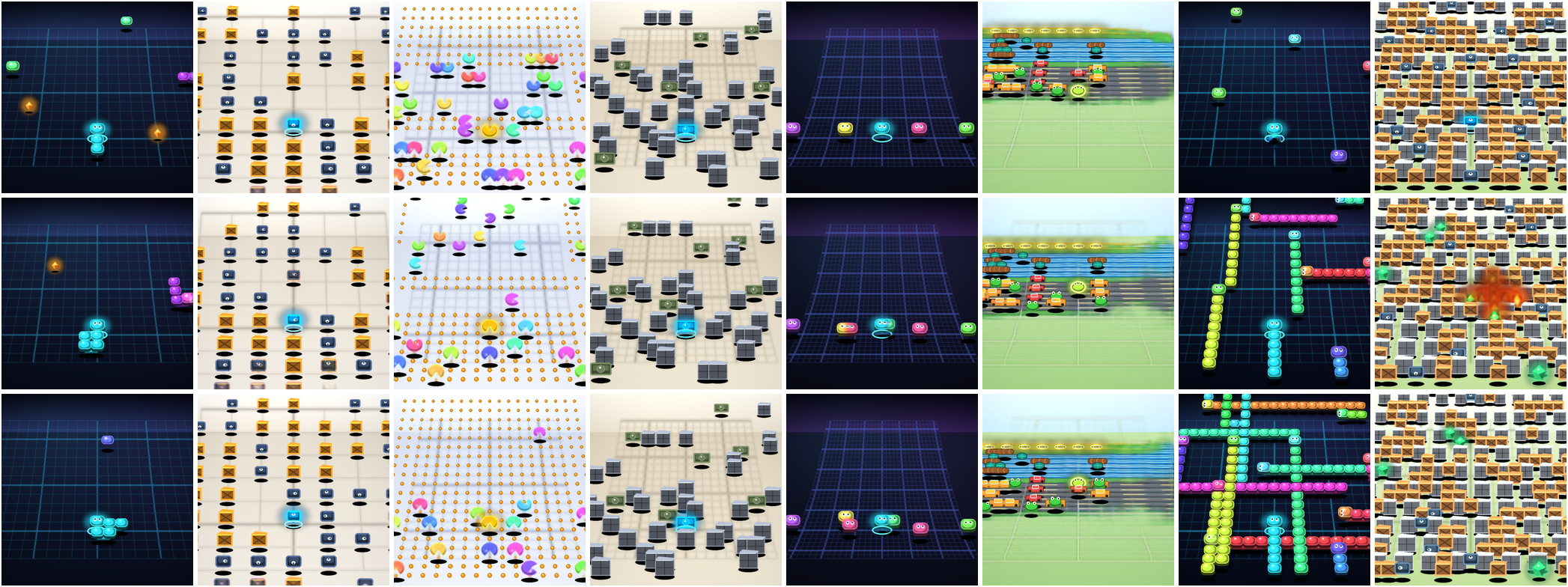}
\caption{State-conditioned renders across eight games. Columns show Snake,
Crate Pusher, Pac-Man, Tank Battle, Lunar Touchdown, Frogger, Tron, and
Bomberman. Rows show different timesteps.}
\label{fig:renders-n1024}
\end{figure}

\begin{table}[t]
\centering
\small
\caption{Held-out reconstruction quality at $256\times256$ resolution.
Object PSNR is measured on entity pixels.}
\label{tab:renderers}
\begin{tabular*}{\columnwidth}{@{\extracolsep{\fill}}lrrr@{}}
\toprule
Game & PSNR (dB) $\uparrow$ & Object PSNR $\uparrow$ & SSIM $\uparrow$ \\
\midrule
Snake & 38.82 & 29.44 & 0.996 \\
Crate Pusher & 28.50 & 29.64 & 0.987 \\
Pac-Man & 33.95 & 30.50 & 0.987 \\
Tank Battle & 32.21 & 29.93 & 0.988 \\
Lunar Touchdown & 39.83 & 31.49 & 0.995 \\
Frogger & 40.24 & 37.26 & 0.995 \\
Bomberman & 27.82 & 27.58 & 0.969 \\
Tron & 23.72 & 23.20 & 0.910 \\
\bottomrule
\end{tabular*}
\end{table}

\begin{figure}[t]
\centering
\includegraphics[width=1\textwidth]{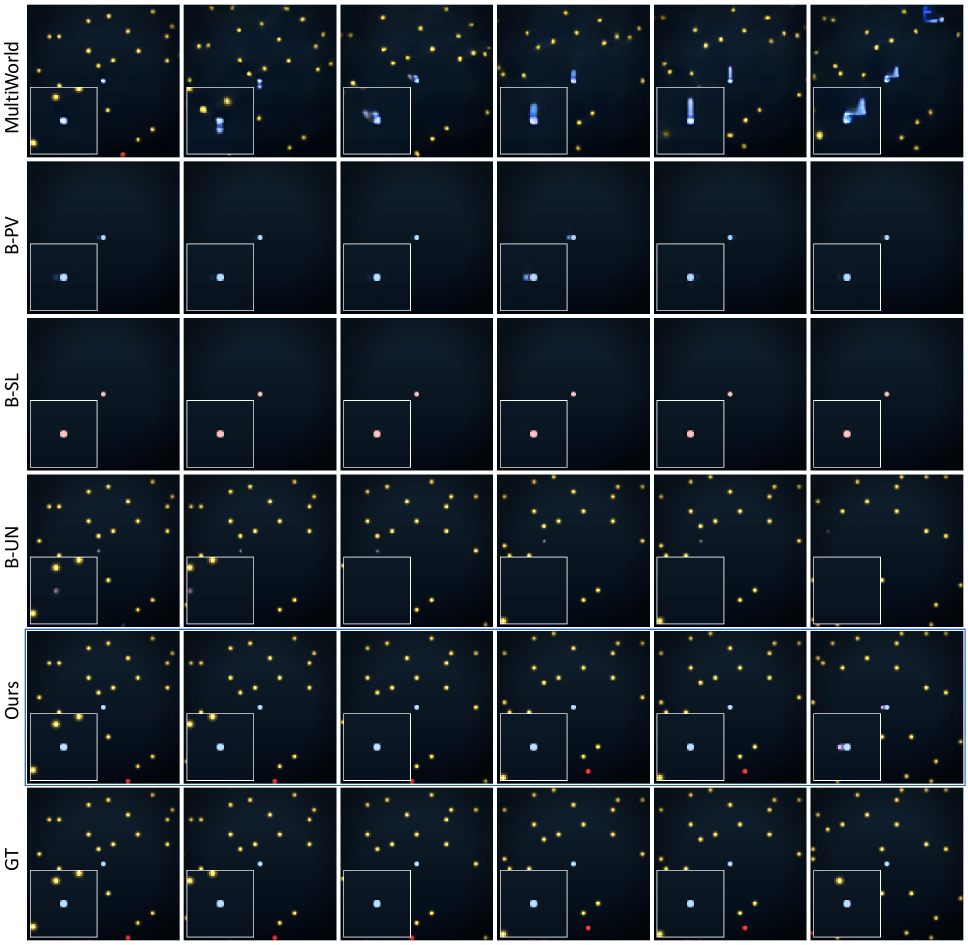}
\caption{Frames from a held-out matched Snake rollout. Rows compare
MultiWorld, B-PV, B-SL, B-UN, \methodname{}, and the ground truth.}
\label{fig:matched-qual}
\end{figure}

\subsection{Setup}
\label{sec:setup}

The matched benchmark, using identical held-out episodes and synchronized cameras, evaluates every method from the same initial world at
$128\times128$ resolution over 128 ticks. Training episodes, joint actions, camera paths, and
optimization budgets remain fixed, so the comparison tests how each recurrent
carrier preserves shared game information. The appendix reports the full protocols.

\subsection{Training details}
\label{sec:training}

The matched Logic Engine uses AdamW with learning rate $2\times10^{-4}$,
weight decay $10^{-4}$, batch size 8, gradient clipping at 1.0, bfloat16
arithmetic, and 20,000 one-step teacher-forced updates.
Autoregressive evaluation decodes greedily and feeds back every
predicted token without native transition calls, state replacement, or
gameplay-rule repair.

\subsection{Baselines}
\label{sec:baselines}

The matched benchmark compares recurrent carriers under identical data and
budgets. MultiWorld \citep{multiworld2026} is a publicly released multiplayer
world model that jointly generates coordinated views, and we retrain its
official code on the benchmark data. B-PV advances each client view with an
independent video predictor. B-SL advances one shared visual latent from which
every view is decoded. B-UN keeps the rest of the \methodname{} pipeline and
replaces the typed carrier with a dense joint state predictor.
The baseline designs are summarized in \Cref{tab:architecture-comparison}.
All baselines receive an equivalent representation of the same reconstructed spawn information.

\subsection{Metrics}
\label{sec:metrics}

We score the matched rollouts by perceptual quality, state recovery, agreement
between simultaneous views, and structural validity. LPIPS measures
perceptual quality \citep{zhang2018unreasonable}. A frozen parser measures
overall state recovery, entity counts, positions, and event F1 from each
rendered frame. X-view records disagreement between simultaneous views.
Invalid is the fraction of recovered states that break the game structure or its consistency.

Logic-only studies compare the predicted typed state directly with the engine
state using field accuracy, full-state exact match, and structural validity.
State-Referenced Semantic Correctness, or SRSC, provides complementary vision
and language scores for logic, rendering, and the complete prediction
\citep{zheng2023judging,hu2023tifa}. The appendix gives the complete protocols.

\subsection{Main results}
\label{sec:main-results}

The matched Snake results are shown in \Cref{tab:matched}.
\methodname{} leads six of the seven metrics. Its LPIPS of 0.098 is the lowest
perceptual error among all methods, compared with 0.123 for the next best
(B-UN) and 0.277 for MultiWorld. Parser recovery reaches 0.764, more than
five times the highest video-based baseline (B-PV at 0.128).
State-level consistency is guaranteed by construction because every rendered
view decodes from the same predicted typed state.  In the matched benchmark
this yields a measured X-view disagreement of 0.000, confirming that the
learned Rendering Engine does not introduce cross-view inconsistencies. B-UN achieves competitive LPIPS yet its parser recovery
falls to zero, confirming that pixel-level agreement does not guarantee a
recoverable world state. \methodname{} preserves entity counts, positions, and
interaction outcomes while every requested camera inherits one consistent
prediction.

Learned renders across eight games are shown in \Cref{fig:renders-n1024},
and \Cref{tab:renderers} reports their held-out reconstruction quality.
The same Logic Engine and Rendering Engine architecture, instantiated per game
with game-specific weights, runs on every game through its declarative schema. PSNR ranges from 23.72\,dB for Tron to 40.24\,dB for Frogger, with five of the eight games exceeding 32\,dB. Object PSNR, which
measures pixel accuracy on entity regions, stays above 27.5\,dB across most
games. These results indicate the typed-state interface supports diverse
visual themes without per-game architecture changes beyond tokenizers,
embeddings, and model weights.

The same transition architecture also advances worlds with 1{,}024 simulated
player entities for 10{,}000 recurrent ticks. Because the learned transition
runs once per tick, the cost of advancing the world is independent of the
number of rendered views. State distribution and requested rendering remain
separate workloads. The appendix reports per-horizon results and measured
throughput.

\begin{table}[t]
\centering
\small
\caption{Client prediction during missing server updates on Snake.}
\label{tab:client-prediction}
\begin{tabular*}{\columnwidth}{@{\extracolsep{\fill}}lcccc@{}}
\toprule
& $k{=}1$ & $k{=}2$ & $k{=}4$ & $k{=}8$ \\
\midrule
In-view agreement, oracle joint inputs & 1.000 & 1.000 & 1.000 & 1.000 \\
In-view agreement, local actions only & 0.815 & 0.652 & 0.604 & 0.429 \\
Local-avatar displacement (cells) & 0.00 & 0.00 & 0.00 & 0.00 \\
\bottomrule
\end{tabular*}
\end{table}

\subsection{Qualitative results}
\label{sec:qualitative-results}

The held-out rollout in \Cref{fig:matched-qual} matches the quantitative
comparison.
MultiWorld develops smeared or misplaced entities, while B-PV and B-SL retain
little beyond the selected player. \methodname{} keeps the food field, snake
geometry, and interaction outcomes aligned with the reference through the
128-tick rollout.

\subsection{Ablations}
\label{sec:ablations}

We test the typed carrier and the separation between dynamics and rendering.

\paragraph{Typed carrier.}
Replacing the typed carrier with the dense joint predictor reduces parser
recovery to zero and nearly eliminates recoverable interaction events (B-UN ablation), even
though LPIPS remains competitive at 0.123. The dense grid does not preserve
entity identity across ticks because the same grid cell must encode a
different entity at each step, and the U-Net lacks an explicit mechanism
for tracking which object is which between frames. The typed carrier
connects visual agreement to a recoverable world state by keeping each
entity record separate. Further logic-only comparisons appear
in the appendix.

\paragraph{Error localization.}
The explicit state interface lets us measure dynamics and rendering
separately. The SRSC logic score tracks state degradation, while renderer
fidelity remains stable when the predicted state is replaced by the reference
state. This separation locates an error in world evolution or appearance
synthesis. The complete component analysis appears in the appendix.

\subsection{Client prediction}
\label{sec:client-prediction}

When server updates stall, a client advances the latest authoritative state
with the same Logic Engine. It supplies its own actions and uses no-op actions
for the other players. \Cref{tab:client-prediction} reports in-view
agreement, defined as the multiset intersection-over-union of visible
(kind,~x,~y) objects in the predicted and server states, after one,
two, four, and eight missing server updates.

The local avatar remains correct for every evaluated stall length, so the next
server update requires no position correction for that player. Agreement for
other visible objects decreases as their unknown actions accumulate. With
oracle joint inputs, the client rollout matches the server for all eight ticks,
which attributes the difference to missing remote actions rather than the
learned transition.

\section{Conclusion}
\label{sec:conclusion}

\methodname{} brings the authoritative-server architecture of multiplayer
games to learned world models. A shared predicted state keeps simultaneous
cameras consistent by construction and separates world-transition cost from
the number of rendered views. In our experiments, the same compact Logic
Engine and Rendering Engine support several games through declarative schemas,
long structurally valid rollouts, client-side prediction without local-player
displacement, and worlds with 1{,}024 simulated player entities. The
typed-state interface locates errors in dynamics or rendering before a frame
is ever generated, and the schema mechanism lets new games use the same
architecture without modification. These results demonstrate that disentangling world dynamics from view
synthesis provides a practical foundation for large learned multiplayer
worlds. Extending the same principle to 3D environments and richer entity
interactions is a natural direction for future work.

\section*{Acknowledgments}
This study was carried out using the TSUBAME4.0 supercomputer at Institute of Science Tokyo.

\clearpage

\bibliographystyle{plainnat}
\bibliography{references}

\clearpage
\beginappendix
\setcounter{figure}{0}
\setcounter{table}{0}
\renewcommand{\thefigure}{\thesection.\arabic{figure}}
\renewcommand{\thetable}{\thesection.\arabic{table}}
\graphicspath{{figures/}}

\section{\methodname{} in detail}
\label{app:details}
\subsection{A real game schema}
\label{supp:schema-example}

All game-specific configuration in \methodname{} enters through one
declarative file. The listing below is a lightly abridged version of the
schema used for the population-scale Snake results (rollouts with $N{=}1024$ and
1{,}024-view walls). The matched benchmark uses a separate compact codec
described in Appendix~\ref{supp:protocol}. It declares the
typed tables of the world (1{,}024 snakes and 4{,}096 spatially bucketed food
rows), the spatial anchor of each record, and the neighborhood layers written
into the context window $c_t^i$ (Eq.~\ref{eq:record-tokenizer}). The Logic
Engine receives no transition rules, rewards, or game code. A new game supplies
a schema of this form and recorded trajectories.

\begin{lstlisting}
schema_version: snake-n1024-factorized-v4
game: snake
engine_state_schema: snake-1
representation_boundary: serde_serialization_only
generated_from:
  state_type: engine_snake::state::SnakeWorldState
  state_source: game-engine/crates/engine-snake/src/state.rs
  agreement: state_json_to_typed_rows_roundtrip_exact

factorization:
  constants:
    arena: [512, 512]
    viewport_cells: 31
  globals:
    row_kind: global
    fields: [tick]
  tables:
    - state_field: snakes          # one row per player entity
      row_kind: snake
      rows: 1024
      matching: index
      action: joint_index          # receives that player's action token
      anchor: body.0               # records are spatially anchored at the head
    - state_field: food            # world-owned entities, spatially bucketed
      row_kind: food_bucket
      rows: 4096
      matching: index
      canonical: sorted_unique     # decoded as a canonical sorted set
      anchor: $index
      layout:
        type: spatial_buckets
        width: 512
        height: 512
        bucket_width: 8
        bucket_height: 8
  spatial_context:                 # what each record sees around its anchor
    arena_field: arena
    radius: 4
    layers:
      - {source_field: food,   points: $item,  value: {fixed: food}}
      - {source_field: snakes, points: body,   value: {fixed: body}}
      - source_field: snakes
        points: body.0
        indexed_action: true       # neighbors' pending actions are visible
        value:
          template: head_{heading_x}_{heading_y}_a{action}
# ... field vocabularies, bounded body bucketing, and canonical-order
# declarations follow (157 lines total in the repository).
\end{lstlisting}

\subsection{Additional details of \methodname{}}
\label{supp:method-details}

\methodname{} uses the shared world state as its recurrent object. The Logic
Engine advances this state once, and the Rendering Engine answers camera
queries from the result. Algorithm~\ref{alg:supp_shared_world_step} describes
the complete learned update. Here,
$\mathcal{M}_{\Sigma_g}$ denotes the field masks derived from the schema and
$P_c$ denotes the deterministic projection local to camera $c$.

\begin{algorithm}[H]
\small
\caption{One learned shared-world update.}
\label{alg:supp_shared_world_step}
\begin{algorithmic}[1]
\REQUIRE Schema $\Sigma_g$, recurrent state \sharedstate{t}, version $v_t$
\REQUIRE Joint action \jointaction{t}, declared input $e_t$, cameras $\mathcal{C}_{t+1}$
\ENSURE State \sharedstate{t+1}, version $v_{t+1}$, views $\{\hat I_{t+1}^{c}\}_{c\in\mathcal{C}_{t+1}}$
\STATE $\mathcal{R}_t \gets \operatorname{Records}_{\Sigma_g}(\sharedstate{t})$
\FORALL{$\rho_t^i \in \mathcal{R}_t$}
    \STATE $c_t^i \gets \operatorname{Context}_{\Sigma_g}(\rho_t^i,\mathcal{R}_t)$
    \STATE $u_t^i \gets T_{\Sigma_g}(\rho_t^i,c_t^i,\jointaction{t}^{\pi(i)},e_t^i)$
    \STATE $\hat{\rho}_{t+1}^i \gets \operatorname{Decode}_{\theta}(u_t^i,\mathcal{M}_{\Sigma_g})$
\ENDFOR
\STATE $\sharedstate{t+1} \gets \mathcal{A}_{\Sigma_g}(\{\hat{\rho}_{t+1}^i\}_{i=1}^{K_g})$
\STATE $v_{t+1} \gets v_t+1$
\FORALL{$c \in \mathcal{C}_{t+1}$}
    \STATE $x_{t+1}^{c} \gets P_c(\sharedstate{t+1})$
    \STATE $\hat I_{t+1}^{c} \gets \renderermodel(x_{t+1}^{c})$
\ENDFOR
\RETURN $\sharedstate{t+1},v_{t+1},\{\hat I_{t+1}^{c}\}_{c\in\mathcal{C}_{t+1}}$
\end{algorithmic}
\end{algorithm}

The record loop builds every prefix from the same current state and evaluates
them as one batch. Assembly begins after all next records have been decoded.
The camera loop then reads the assembled state without changing it. Each tick
therefore contains one world transition, while rendering cost grows with the
requested camera set. Every view uses the same predicted world with its own
camera and player conditioning.

\paragraph{Typed world state and game schemas.}\label{supp:typed-state}
A schema specifies the entities that can occur in a game, the fields attached
to each entity, and the legal domain of each field. At tick $t$, the resulting
state is an ordered collection of typed records
\begin{equation}
s_t=\bigl(\rho_t^1,\ldots,\rho_t^{K_g}\bigr),
\qquad
\rho_t^i=(\text{kind}_i,\text{fields}_i),
\label{eq:supp-typed-state}
\end{equation}
where $K_g$ is determined by the schema and scale of game $g$. The schema
defines the representation. It identifies which values form a state, but it does
not specify how a snake grows, how a crate moves, or how a collision is
resolved. Those outcomes are learned from recorded transitions.

Canonical serialization makes Eq.~\eqref{eq:supp-typed-state} unambiguous.
Record kinds follow a fixed schema order, and records within an entity table
are indexed by persistent slots or a declared canonical key. Booleans,
categorical attributes, and integer coordinates use disjoint typed token
ranges. Optional fields receive an explicit present or absent token. Bounded
lists begin with their length and end in canonical padding. Spatial buckets
and other variable-size collections are sorted and deduplicated before
serialization. An unused entity slot is therefore an inactive typed record,
not a missing position in the sequence. Decoding reverses these steps exactly
and restores the full state consumed by the next rollout step.

The matched Snake codec illustrates how a tick is flattened into tokens.
The model input is a flat integer sequence assembled from four segments:
the encoded state, player actions, exogenous inputs, and an output marker.
Every field occupies a fixed position range, so the decoder mask
(Eq.~\eqref{eq:supp-logic-loss}) constrains each output position to its
declared vocabulary. Below is the prefix structure for a two-player episode at tick~142:

\begin{lstlisting}
# Matched Snake input prefix (447 tokens). Each [value] is an integer token;
# the ranges (NIBBLE, CELL, etc.) are defined by the schema.

[0]                          # BOS
[1]                          # STATE marker

# --- state segment (421 tokens) ---
[5, 5, 13, 19]               # tick = 142, as 4 hex nibbles (base 5..20)
[CELL, CELL, ..., CELL]      # 64 sorted food cell coordinates
[11]                         # player_count = 2

# Player slot 0: body length 3, heading east, alive
[BODY_LEN]                   # body_length = 3
[CELL, CELL, CELL]           # body cells (23,45), (22,45), (21,45)
[PAD] x37                          # padding to MAX_BODY_LENGTH = 40
[HEADING]                    # heading = east
[ALIVE]                      # alive = true
[PAD]                        # dead_at = None (padding sentinel)

# Player slot 1: body length 1, heading west, alive
[BODY_LEN]                   # body_length = 1
[CELL]                       # body cell
[PAD] x39                          # padding
[HEADING]                    # heading = west
[ALIVE]                      # alive = true
[PAD]                        # dead_at = None

# Player slots 2-7: inactive (every field is a padding token)
[PAD] x (44 x 6)                 # 44 padding tokens for each of six inactive player slots; 4+64+1+8x44=421

# --- action segment (8 tokens) ---
[2]                          # ACTION marker
[4, 2, 0, 0, 0, 0, 0, 0]    # player 0 = right, player 1 = up, rest no-op

# --- exogenous inputs (13 tokens) ---
[3]                          # EXOGENOUS marker
[5, 5, 13, 19]               # tick = 142
[7]                          # spawn_count = 2
[CELL, CELL]                 # spawn at (10,22) and (43,7)
[PAD] x6                           # padding to MAX_FOOD_SPAWNS = 8

[4]                          # OUTPUT marker (the model decodes the next state here)
\end{lstlisting}

The decoder receives the entire prefix and predicts the next state tokens
autoregressively after the \texttt{OUTPUT} marker. The schema-derived mask
ensures that each field decodes to its declared type, for example the tick
position can only produce nibble-range values, heading positions produce only
the four cardinal directions plus a base token, and food coordinates are
constrained to the arena and forced into sorted order by the dynamic selector.
Body cells within one player are constrained to be unique within that body.
These constraints make every decoded state structurally valid by construction.
The model only needs to learn the transition dynamics, not the representation
format.

Table~\ref{tab:supp-state-representations} gives the layouts used in the main
experiments. The compact matched Snake benchmark serializes one complete small
state into a single sequence. The population-scale codecs factorize larger
states into short record sequences and batch those records. Both use the same
typed-state interface and train separate models. Across games, the experiment
reuses the Transformer architecture and training recipe with game-specific
weights.

\begin{table*}[t]
\centering
\small
\setlength{\tabcolsep}{2pt}
\renewcommand{\arraystretch}{0.9}
\caption{Population-scale typed state layouts. The matched Snake benchmark uses a separate compact codec (see Appendix~
\ref{supp:protocol}).}
\label{tab:supp-state-representations}
\begin{tabular*}{\textwidth}{@{\extracolsep{\fill}}y{30}y{65}y{90}y{60}y{65}y{118}@{}}
\toprule
Regime & Record kinds & Representative fields & Records/world & Max. tokens/record & Local context \\
\midrule
Matched Snake & global, snake, food & tick, body, heading, alive, food cells & 1 & 867 & complete-state prefix \\
Snake, $\ntick{=}1024$ & global, snake, food bucket & body, heading, death anchor, bucket entries & 5,121 & 513 & $9\times9$ anchor grid \\
Crate Pusher, $\ntick{=}1024$ & global, wall, goal, crate, pusher & location, occupancy, goal, heading & 2,583 & 124 & $9\times9$ anchor grid \\
Pac-Man, $\ntick{=}1024$ & global, pellet chunk, pacman, ghost & position, direction, life, pellets & 2,049 & 2,590 & $3\times3$ grid, chunk-local agents$^{*}$ \\
\bottomrule
\end{tabular*}
\vspace{2pt}
\parbox{0.98\textwidth}{\footnotesize $^{*}$A pellet-chunk row also receives
up to 64 Pac-Man state and action records from its $16\times16$ chunk and
one-cell halo.}
\end{table*}

For record-wise prediction, local context is computed from the current state
before any next-state field is decoded. Population-scale Snake and Crate
Pusher use a schema-defined $9\times9$, radius-four grid centered on the
record anchor. The selected Pac-Man codec instead supplies every row with a
$3\times3$, radius-one anchor grid. Each $16\times16$ pellet-chunk row also
receives up to 64 Pac-Man state and action records whose positions fall in that
chunk enlarged by one cell on each boundary. Global, Pac-Man, and ghost rows
do not receive this auxiliary list. These current-state contexts support local
interactions without allowing a record to read another record's unknown
next-state label. Matched Snake differs by predicting the complete state from
one full-state prefix rather than using a separate spatial patch.

Exogenous inputs carry randomness that is supplied to the rollout rather than
predicted as an interaction outcome. In the population-scale experiments they
are engine-recorded tokens, such as prospective food spawns, and are replayed
with the action stream during evaluation.

\paragraph{Logic Engine.}\label{supp:logic-engine}
The Logic Engine is a decoder-only causal Transformer with width 256, six
layers, eight attention heads, an MLP expansion ratio of four, tied token
embeddings, learned sequence positions, and coordinate embeddings. The
compact matched model has 5.66 million parameters and a configured maximum
length of 1,024. Record-wise models use the same width, depth, and attention
configuration. Their embedding tables and maximum length follow each schema,
so parameter counts vary with the vocabulary.

For prediction unit $i$, the codec forms a prefix
\begin{equation}
u_t^i=T_{\Sigma_g}(\rho_t^i,c_t^i,a_t^{\pi(i)},e_t^i),
\label{eq:supp-record-prefix}
\end{equation}
where $c_t^i$ is the current spatial context, $a_t^{\pi(i)}$ is the associated
player action when the record has an owner, and $e_t^i$ is the relevant share
of the exogenous input. Target tokens are appended autoregressively. Training
minimizes token cross entropy under teacher forcing,
\begin{equation}
\mathcal{L}_{\mathrm{logic}}
=-\sum_{i=1}^{K_g}\sum_j
\log p_\theta\bigl(y_{t+1}^{i,j}\mid u_t^i,y_{t+1}^{i,<j}\bigr).
\label{eq:supp-logic-loss}
\end{equation}
At inference, the model decodes every field greedily, assembles the predicted
records into $\sharedstate{t+1}$, and feeds that prediction into the next step.

A schema-derived mask restricts each output position to its field vocabulary.
Dynamic selectors enforce canonical list padding, sorted unique collections,
and per-list uniqueness where these are part of the serialization contract.
Movement, collision, growth, collection, death, and reward remain learned
outcomes rather than selector rules. The direct state metrics in
Appendix~\ref{supp:matched-extended} therefore measure semantic transition
accuracy separately from structural decodability.

Algorithm~\ref{alg:supp_shared_world_step} also fixes the rollout ordering.
Every prefix is built from one current predicted state. All next records are
decoded before assembly, and assembly precedes publication and rendering. The
same assembled prediction is the recurrent input, synchronization object,
target for direct evaluation, and source for all requested images.

\paragraph{Rendering Engine.}\label{supp:rendering-engine}
Rendering starts with a camera-local projection $P_i(\sharedstate{t})$. The
projection rasterizes visible occupancy, semantic role, ownership and selected
player identity, depth or range, and camera geometry into a 16-channel tensor.
The camera is anchored to the selected entity and parameterized by heading,
pitch, and a bounded field of view. Projection occurs independently for each
requested camera, but every projection reads the same complete predicted
state.

The native-resolution Rendering Engine is a geometry-aware residual U-Net. A
64-channel stem processes the 16 semantic channels together with coordinate,
semantic-gradient, and depth-gradient features. Three downsampling stages,
five dilated residual blocks at the bottleneck, and three skip-connected
upsampling stages produce a sigmoid RGB output. The 2D and 2.5D scale studies
use the same state and camera contract with a lighter shared U-Net whose width is
selected per game. Resolutions are $128\times128$ in the matched benchmark,
$96\times96$ in the six-renderer quantitative matrix, and $256\times256$ in
the native-resolution comparison across games.

Renderer training pairs a ground-truth typed state and recorded camera with
an offline target image. The loss combines object-weighted Charbonnier
reconstruction, multiscale $L_1$, image-gradient alignment, SSIM, and color
saturation terms. No perceptual or adversarial network is used. At inference,
the renderer receives neither target pixels nor RGB history. A predicted-state
rollout therefore exposes logic error to every camera, while a ground-truth
state rendered by the same network isolates image-synthesis error.

\paragraph{Client-side prediction.}\label{supp:client-method}
A client caches the latest versioned authoritative state. If server updates
are temporarily missing, it initializes the same Logic Engine from that state
and advances the full typed world locally. The full-input reference supplies
the complete recorded joint-input stream. The deployable local-actions
variant supplies the action of the evaluated player, assigns no-op actions to the
other players, and uses empty future-spawn lists. Its predicted state becomes
both the next recurrent input and the source of the local view.

When a new authoritative version arrives, the client replaces the speculative
state with that version and resumes normal rendering. Reconciliation uses the
versioned typed state directly, without a separate correction network or a
hidden RGB latent. Appendix~\ref{supp:large-scale-async} measures how agreement
changes as the update gap grows under full-state prediction. Transport
scheduling and neighborhood prediction belong to the surrounding systems
layer.

\setcounter{figure}{0}\setcounter{table}{0}
\section{The matched multiplayer benchmark}
\label{app:matched}
\subsection{Experimental protocol}
\label{supp:protocol}

The matched benchmark tests whether a recurrent carrier preserves one
recoverable world across two views. Direct state evaluation measures this
property before rendering. Population-scale and cross-game studies test the
same state and camera decomposition with different world sizes, camera counts,
and visual domains.

\paragraph{Datasets and splits.}\label{supp:data}
The matched benchmark contains multi-agent Snake trajectories on a
$48\times48$ arena. The data are separated by episode. The training split has
128 episodes at each of $\ntick{}\in\{2,4,8\}$, with 192 transitions per
complete episode. Validation and test contain 16 and 32 episodes per
population, respectively, with 160 transitions per episode. Direct logic
evaluation uses the first eight complete validation episodes at every
population, for 24 episodes in total. The end-to-end benchmark uses the
$\ntick{=}2$ subset required by the two-view game interface of
MultiWorld~\citep{multiworld2026}. Of the 128 training episodes, 126 provide
the required 129 synchronized frames. The other two end earlier. Formal image
evaluation uses the first eight complete test episodes, two synchronized
camera views per episode, and a 128-tick rollout at $128\times128$.

Every method starts from the same recorded state and initial images, follows
the same recorded joint actions, and is scored against the same world and
camera trajectories. Cameras are player centered with a 31-cell viewport.
Their paths are fixed from the recorded reference state rather than recentered
from the prediction of a method, so all methods are compared at identical view
locations. The selected trajectories include food collection, growth,
collisions, deaths, and interactions between nearby players.

The compact cross-game state study changes only the codec and trajectories.
Breakout provides 3,466 training and 855 validation transitions, while Tile
Merger provides 5,729 and 1,916. Evaluation includes every eligible
autoregressive window, yielding 220 Breakout and 900 Tile Merger windows
through $H=128$. The population-scale study uses separately generated Snake,
Crate Pusher, and Pac-Man trajectories with 1,024 controlled entity slots and
native exogenous tokens. Long rollouts start from held-out recorded states and
reuse their action and exogenous streams for as long as a reference remains
available.

Renderer datasets are split by source state trajectory. Each example contains
one complete state, one camera tensor, and one target RGB frame. Ground-truth
states supervise renderer training. Evaluation uses held-out states. For
end-to-end figures, the renderer receives the Logic Engine's predicted state
instead of the ground-truth state, with no change to the renderer weights.

\paragraph{Baselines and fair comparison.}\label{supp:baselines}
The matched benchmark compares five recurrent carriers. MultiWorld
\citep{multiworld2026} is the publicly released coupled-view model, retrained
from its official implementation on the matched Snake corpus. B-PV predicts
each view independently and therefore has
no shared recurrent object. B-SL advances one visual latent and decodes both
views from it. B-UN replaces the typed carrier with a dense joint state U-Net
while retaining the learned state-to-view path. \methodname{} advances the
canonical typed state and renders both cameras from that prediction.

Table~\ref{tab:supp-model-configs} records the settings needed to interpret
the comparison. All methods see the same training episodes, actions, camera
paths, and 128-tick test protocol. The internal baselines and \methodname{}
use the same 20,000-update budget. MultiWorld follows the official 35-epoch
optimization schedule on the same matched corpus. Parameter counts refer to
the recurrent predictor when logic and rendering are separate. The MultiWorld
entry reports trainable parameters in the adapted official model.

\begin{table*}[t]
\centering
\small
\setlength{\tabcolsep}{3pt}
\renewcommand{\arraystretch}{0.9}
\caption{Matched benchmark configurations.}
\label{tab:supp-model-configs}
\begin{tabular*}{\textwidth}{@{\extracolsep{\fill}}y{46}y{80}y{58}y{48}y{55}y{60}y{75}@{}}
\toprule
Method & Recurrent carrier & Predictor parameters & Training schedule & Training context & Output & Test-time recurrence \\
\midrule
\methodname{} & typed token Transformer & 5.66M & 20k updates & one transition & canonical state $+$ RGB & greedy predicted state \\
MultiWorld & coupled multiview diffusion & 5.60B trainable & 35 epochs & official video window & two joint RGB views & generated visual state \\
B-PV & independent video predictors & 8.13M & 20k updates & 8 frames & one RGB per view & generated frames \\
B-SL & shared visual latent & 12.72M & 20k updates & 8 frames & two RGB views & generated latent \\
B-UN & dense joint-state U-Net & 3.69M & 20k updates & four transitions & dense state $+$ RGB & raw predicted state \\
\bottomrule
\end{tabular*}
\end{table*}

B-UN is the controlled dense-state diagnostic for the representation question.
It shares the matched trajectories, split, update budget, initialization,
horizons, and metric code, but carries the world in a dense joint map. The
comparison therefore tests whether a dense shared carrier can preserve sparse
entity identity and position under recurrence.

\paragraph{Evaluation metrics.}\label{supp:metrics}\ 
Image-space metrics are computed from lossless RGB outputs. LPIPS measures
perceptual distance between prediction and reference \citep{zhang2018unreasonable}. A frozen Snake parser
maps each frame to a 31-by-31 grid of background, food, and entity labels plus
the selected-player mask. Parser is the foreground-position F1 of this
recovered grid. Count is exact entity-count accuracy per view, and Position is
foreground-position recall. Event F1 compares foreground changes between
successive overlapping view regions, so collection, growth, entry, and exit
must occur at the correct location and time.

The parser is evaluated on the native ground-truth frames before it is used on
model output. Semantic evaluation is enabled only when its ground-truth state
recovery is at least 0.99. The frozen parser passes this criterion on the
matched test set. This check separates model error from parser error on the
target rendering style.

Cross-view disagreement aligns overlapping camera regions in world
coordinates using the recorded camera origins. It counts parsed foreground
cells whose semantic class differs between two simultaneous views, divided by
the union of foreground cells in the overlap. The metric recovers content
independently from each generated image before comparison. X-view
measures disagreement, while Parser, Count, Position, and Event F1
provide the complementary comparison with the reference world.

Invalid is the fraction of parsed ticks that violate structural constraints
such as a missing selected entity or an inconsistent recovered world. We
interpret it jointly with Parser, Count, and Position so that structural
validity and recoverable world content remain distinct measurements.

Direct logic evaluation compares predicted typed fields with the recorded
engine state before rendering. Semantic field accuracy averages canonical
field correctness, Active restricts the score to populated fields, Position
tracks entity anchors, Count tracks the number of decoded entities, and Full
exact requires the complete canonical state to match. Structural validity
checks only the declared representation constraints.

State-Referenced Semantic Correctness, or SRSC, is an auxiliary RGB diagnostic
inspired by calibrated vision and language evaluation
\citep{zheng2023judging,hu2023tifa}.  It uses four paired comparisons, each
written as an explicit (reference, candidate) tuple in the protocol details
(\S\ref{supp:srsc}).  The Logic axis compares a teacher rendering of the
ground-truth state with a teacher rendering of the predicted state.  The
Renderer-GT axis compares teacher and learned renderings of the same
ground-truth state.  The Renderer-Pred axis compares teacher and learned
renderings of the same predicted state.  The End-to-end axis compares a
teacher rendering of the ground-truth state with a learned rendering of the
prediction.  The frozen judge is Qwen3.5-27B with deterministic greedy
decoding (temperature~0) and a pinned seed, using the mean 0--100 score
as the reported metric.  Its logic-axis score correlates with the
programmatic state metric ($r=0.962$).  SRSC localizes errors across the
pipeline, while the deterministic metrics provide the quantitative
conclusions.

The population-scale Logic Engines use the same 256-wide, six-layer,
eight-head architecture and AdamW recipe. Snake and Crate Pusher are evaluated
at 20,000 updates. The Pac-Man run uses 100,000 updates. Records are shuffled
and batched across worlds, and rare growth, death, and collection rows are
sampled more frequently during training. Generation may split the records of
one state into microbatches for memory, but all records are assembled before
the next recurrent step.

The learned renderers are trained from random initialization with AdamW,
learning rate $2\times10^{-4}$, batch size 8, bfloat16 arithmetic, and
gradient clipping at 1.0. We train the native-resolution Rendering Engine for
6,000 updates and the matched renderer for 20,000 updates. The selected checkpoint
maximizes a held-out combination of PSNR, object PSNR, SSIM, and edge error.
All reported inference frames use the learned checkpoint directly from a
state and camera tensor, without target images, RGB history, or a native-renderer
fallback.

\subsection{Matched Snake codec}
\label{supp:codec}

The matched end-to-end benchmark uses multi-agent Snake on a $48\times48$
arena with at most eight player slots.  Its non-redundant canonical state
contains the tick, 64 sorted food coordinates, and for each player an ordered
body-coordinate list, length, heading, survival bit, and optional death
anchor. The state occupies 421 tokens. The 447-token input prefix
concatenates the current state, all player actions, and a compatibility
payload of up to eight net-added food coordinates. The frozen corpus stores
this compatibility payload through adjacent recorded states because it was
collected before native exogenous-token recording. The payload records whether
a spawn occurred. The matched Snake row evaluates this historical
compatibility interface, while the population-scale studies use prospective
engine-native randomness.

The schema-generated decoder mask limits each field to its declared type and
range. It also keeps food coordinates sorted and unique, prevents repeated
coordinates within a body, and emits canonical padding. Body adjacency, legal
movement, collision, growth, death, food consumption, and other transitions
remain learned rather than encoded in the mask. The state metrics score their
semantic accuracy separately from structural validity. We initialize the
matched renderer randomly and train it for 20{,}000 updates on the frozen Snake
trajectories at $128\times128$.

\subsection{Extended results on matched multiplayer Snake}
\label{supp:matched-extended}

The main text reports aggregate performance after a 128-tick matched rollout.
This section follows performance over time, evaluates the predicted state
before rendering, and pairs states with renderers to separate the stages of the
pipeline.

\begin{figure*}[t]
\centering
\includegraphics[width=0.92\textwidth]{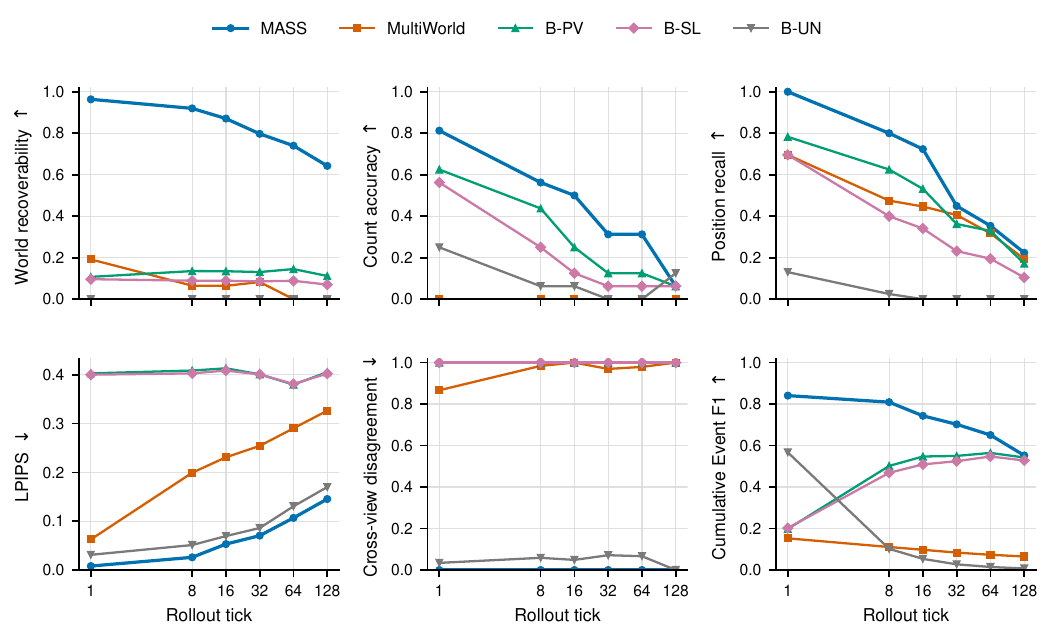}
\caption{Performance over a 128-tick matched Snake rollout. Event F1 is
cumulative through each horizon.}
\label{fig:supp-matched-horizons}
\end{figure*}

\paragraph{Performance over rollout horizon.}\label{supp:horizon-results}
Figure~\ref{fig:supp-matched-horizons} reports the matched comparison over
time. Every point averages the same eight held-out episodes and two
synchronized views used by the main evaluation. The curves separate properties
that an endpoint image score can conflate, including recoverable world content,
agreement between simultaneous views, and image fidelity to the reference.

World recoverability separates the methods from the first predicted tick.
\methodname{} recovers 0.963 of the foreground state at that tick,
while no baseline exceeds 0.19. Its score then declines smoothly to 0.642 at
tick 128. MultiWorld loses most parsable content by tick 8, and B-UN begins
near zero despite a comparatively low LPIPS. Appearance alone therefore does
not reveal whether a recurrent carrier has preserved the entities that define
the game.

Count and position reveal complementary behavior. \methodname{} achieves a position recall of 1.000 at tick 1 and 0.224 at tick 128. B-PV retains some
isolated positions but does not maintain a coherent shared count, while B-UN
quickly loses both. The LPIPS curve is smoother than the state curves for all
methods. This disentangling explains why a visually plausible or low-distance
frame can still represent a different interaction outcome.

Cumulative Event F1 makes the interaction trend explicit. \methodname{} starts
at 0.841 and retains 0.702 through tick 32. The image predictors approach a
similar endpoint Event F1 while their world recoverability remains below 0.15.
Reading Event F1 together with parser and cross-view scores distinguishes event
timing from recoverable entities and agreement between simultaneous views.

Cross-view disagreement tests a different property from fidelity to the
reference. The two observation-entangled baselines disagree on almost every
foreground overlap throughout the rollout. MultiWorld approaches the same
regime after tick 16. B-UN shares a dense state and therefore lowers
disagreement, but its state contains little recoverable world content.
\methodname{} combines agreement with reference fidelity because both cameras
decode one explicit prediction.

\paragraph{Direct logic evaluation.}\label{supp:direct-logic}
The direct state study removes view synthesis and asks whether the recurrent
carrier predicts the entities themselves. Table~\ref{tab:supp-direct-logic}
compares the typed Transformer with three dense-state diagnostics using the
same trajectories, split, 20,000-update budget, initialization, horizons, and
metric code. Evaluation covers 24 validation episodes, eight each at
$\ntick{}\in\{2,4,8\}$. Every model feeds its raw prediction back at the next
step.

\begin{table*}[t]
\centering
\small
\setlength{\tabcolsep}{3pt}
\renewcommand{\arraystretch}{0.9}
\caption{Direct state prediction on the matched Snake validation set.}
\label{tab:supp-direct-logic}
\begin{tabular*}{\textwidth}{@{\extracolsep{\fill}}llrrrrrr@{}}
\toprule
Carrier & $H$ & Semantic $\uparrow$ & Active $\uparrow$ & Position $\uparrow$ & Count $\uparrow$ & Full exact $\uparrow$ & Contradiction $\downarrow$ \\
\midrule
Independent-head CNN & 1 & 95.1 & 82.4 & 2.7 & 100.0 & 0.0 & 100.0 \\
 & 32 & 91.0 & 64.3 & 0.1 & 100.0 & 0.0 & 100.0 \\
 & 128 & 88.9 & 42.7 & 0.0 & 100.0 & 0.0 & 100.0 \\
Joint U-Net, B-UN & 1 & 95.9 & 83.6 & 0.0 & 95.8 & 0.0 & 100.0 \\
 & 32 & 91.8 & 63.9 & 0.0 & 99.9 & 0.0 & 100.0 \\
 & 128 & 89.1 & 42.4 & 0.0 & 100.0 & 0.0 & 100.0 \\
RSSM & 1 & 91.2 & 77.0 & 0.0 & 100.0 & 0.0 & 100.0 \\
 & 32 & 87.2 & 61.5 & 0.0 & 100.0 & 0.0 & 100.0 \\
 & 128 & 84.7 & 40.6 & 0.0 & 100.0 & 0.0 & 100.0 \\
\methodname{} typed tokens & 1 & \textbf{97.9} & \textbf{94.9} & \textbf{99.1} & 100.0 & \textbf{41.7} & \textbf{0.0} \\
 & 32 & 90.8 & \textbf{65.7} & \textbf{90.2} & 100.0 & 0.0 & \textbf{0.0} \\
 & 128 & 86.8 & 34.6 & \textbf{72.3} & 100.0 & 0.0 & \textbf{0.0} \\
\bottomrule
\end{tabular*}
\end{table*}

The aggregate semantic score hides the difference in position accuracy. At one
step, every dense predictor exceeds 91\% semantic accuracy, yet none exceeds
2.7\% position accuracy and every predicted tick is contradictory. The typed
model reaches 99.1\% position accuracy with no contradiction. At $H=32$ it
retains 90.2\% position accuracy with no structural contradiction. Typed
records therefore preserve entity persistence and local interaction outcomes
through recurrent prediction.

At $\ntick{=}8$, the direct metrics separately resolve entity position and
sparse simultaneous events such as multi-food updates. The typed carrier
preserves entity identity and position while making event coordination
measurable directly in state space.

\paragraph{Logic and rendering decomposition.}\label{supp:decomposition}
The explicit state interface supports a four-way comparison without changing
the evaluated trajectory. A native or teacher renderer applied to the
reference state provides the visual reference. Applying the learned renderer
to that state isolates rendering. Applying the teacher renderer to the
predicted state isolates logic in pixel space. Applying the learned renderer
to the prediction gives the full rollout. Table~\ref{tab:supp-srsc} implements
the same decomposition with the SRSC semantic judge.

\begin{table*}[t]
\centering
\small
\setlength{\tabcolsep}{3pt}
\renewcommand{\arraystretch}{0.9}
\caption{SRSC decomposition of logic, rendering, and end-to-end error. Sample counts are reported in Table~\ref{tab:srsc-axes}.}
\label{tab:supp-srsc}
\begin{tabular*}{\textwidth}{@{\extracolsep{\fill}}llrrrrr@{}}
\toprule
Game & Comparison & $H{=}1$ & $H{=}100$ & $H{=}300$ & $H{=}600$ & $H{=}1200$ \\
\midrule
Snake & Logic & 100.0 & 91.2 & 33.8 & 1.2 & 1.2 \\
 & Renderer-Pred & 53.8 & 76.9 & 60.6 & 48.8 & 30.6 \\
 & End-to-end & 55.6 & 51.2 & 31.2 & 7.5 & 2.5 \\
Crate Pusher & Logic & 78.1 & 10.0 & 13.8 & 16.9 & 12.5 \\
 & Renderer-Pred & 78.8 & 7.5 & 0.0 & 12.5 & 5.0 \\
 & End-to-end & 69.4 & 5.0 & 0.0 & 5.0 & 2.5 \\
Pac-Man & Logic & 100.0 & 60.0 & 10.0 & \textsc{N/A} & \textsc{N/A} \\
 & Renderer-Pred & 100.0 & 100.0 & 65.0 & \textsc{N/A} & \textsc{N/A} \\
 & End-to-end & 100.0 & 70.0 & 10.0 & \textsc{N/A} & \textsc{N/A} \\
\midrule
Snake & Renderer-GT & 68.8 & 71.2 & 63.8 & 75.0 & 66.2 \\
Crate Pusher & Renderer-GT & 100.0 & 75.0 & 100.0 & 100.0 & 100.0 \\
\bottomrule
\end{tabular*}
\end{table*}

The reference-state rows isolate renderer fidelity from recurrent state
evolution. Crate Pusher reaches a mean renderer-ceiling score of 95.0 on
ground-truth states. The corresponding Snake score has a mean of 69.0 and a
sample standard deviation of 4.4 points across horizons. These stable
reference-state results show rendering remains consistent when state
quality is fixed. The remaining rows can therefore be interpreted through
recurrent state evolution.

\paragraph{Joint interpretation.}\label{supp:metric-analysis}
The matched results separate visual similarity from recovery of entity state.
B-UN preserves coarse appearance through a dense shared condition but loses
entity position at the first logic step. \methodname{} recurs on the same
typed state used to decode both views, which preserves recoverability and
cross-view agreement together.

\subsection{Logic dynamics}
\label{supp:track-a}

\begin{figure*}[t]
\centering
\includegraphics[width=0.98\textwidth]{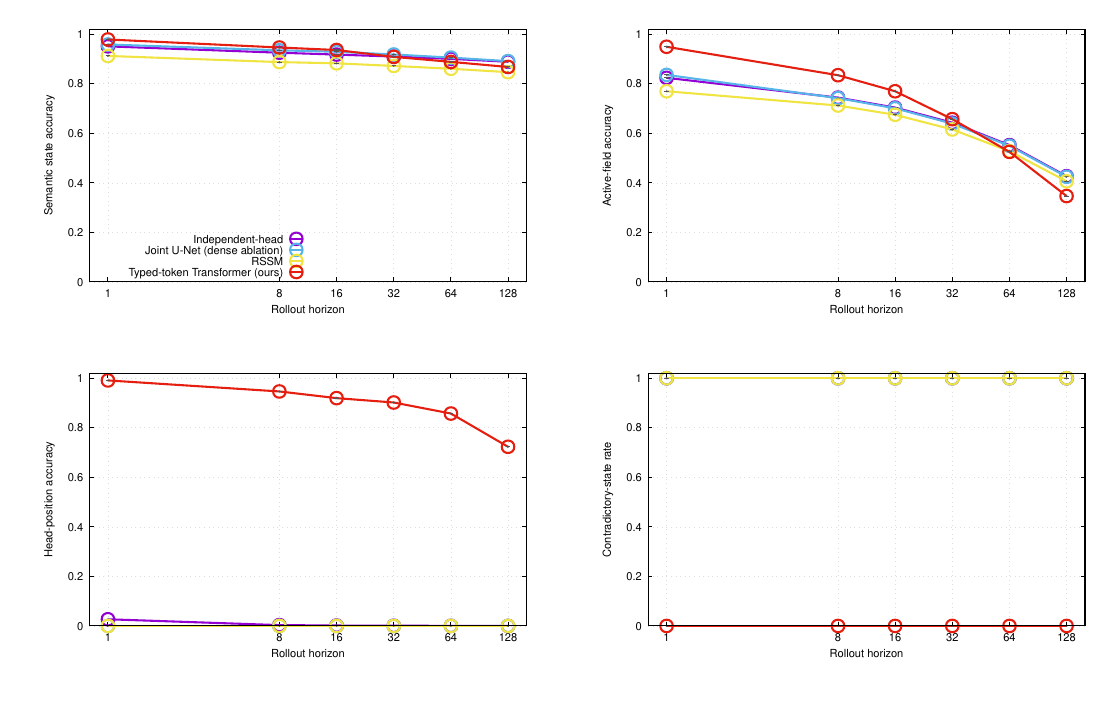}
\caption{Matched Snake validation curves under autoregressive feedback.}
\label{fig:snake-logic-horizon}
\end{figure*}

We compare the typed-token Transformer with three dense-state diagnostics, including a
3.26M independent-head CNN, a 3.86M RSSM with a learned state decoder, and a
3.69M joint U-Net. Every model uses identical trajectories, split,
20{,}000-update budget, rollout initialization, horizons, episodes, and metric
code, and each feeds back its own raw prediction. The typed configuration uses
the compatibility exogenous channel and one-step teacher forcing, while the
dense configurations retain their earlier recorded recipes. The comparison
therefore evaluates the complete carrier configurations. Evaluation covers 24
validation episodes, eight each at $N\in\{2,4,8\}$, and
$H\in\{1,8,16,32,64,128\}$ with a fixed seed. The formal test split remains
sealed for model selection. \Cref{tab:snake-logic-matched,fig:snake-logic-horizon}
summarize the recorded logic results over the rollout horizon.

\begin{table*}[t]
\centering
\footnotesize
\caption{Matched Snake logic results on a fixed validation seed. Values are
percentages except latency.}
\label{tab:snake-logic-matched}
\resizebox{\textwidth}{!}{%
\begin{tabular}{>{\raggedright\arraybackslash}p{52pt}rrrrrrrrrrr}
\toprule
Model & Seeds & $H$ & Semantic $\uparrow$ & Active $\uparrow$ & Position $\uparrow$ & Count $\uparrow$ & Attributes $\uparrow$ & ID switch $\downarrow$ & Contrad. $\downarrow$ & Collapse $\downarrow$ & Latency (ms) $\downarrow$ \\
\midrule
Independent-head & 1 & 1 & 95.1 & 82.4 & 2.7 & 100.0 & 95.1 & 0.0 & 100.0 & 0.0 & 3.28 \\
Independent-head & 1 & 32 & 91.0 & 64.3 & 0.1 & 100.0 & 93.5 & 0.0 & 100.0 & 0.0 & 3.07 \\
Independent-head & 1 & 128 & 88.9 & 42.7 & 0.0 & 100.0 & 91.8 & 0.0 & 100.0 & 0.0 & 3.40 \\
Joint U-Net (dense ablation) & 1 & 1 & 95.9 & 83.6 & 0.0 & 95.8 & 94.9 & 0.0 & 100.0 & 0.0 & 3.92 \\
Joint U-Net (dense ablation) & 1 & 32 & 91.8 & 63.9 & 0.0 & 99.9 & 93.2 & 0.0 & 100.0 & 0.0 & 3.90 \\
Joint U-Net (dense ablation) & 1 & 128 & 89.1 & 42.4 & 0.0 & 100.0 & 91.8 & 0.0 & 100.0 & 0.0 & 3.94 \\
RSSM & 1 & 1 & 91.2 & 77.0 & 0.0 & 100.0 & 86.6 & 0.0 & 100.0 & 0.0 & 3.45 \\
RSSM & 1 & 32 & 87.2 & 61.5 & 0.0 & 100.0 & 83.6 & 0.0 & 100.0 & 0.0 & 3.47 \\
RSSM & 1 & 128 & 84.7 & 40.6 & 0.0 & 100.0 & 82.1 & 0.0 & 100.0 & 0.0 & 3.42 \\
Typed-token Transformer (ours) & 1 & 1 & \textbf{97.9} & \textbf{94.9} & \textbf{99.1} & 100.0 & \textbf{100.0} & 0.0 & \textbf{0.0} & 0.0 & 45.55 \\
Typed-token Transformer (ours) & 1 & 8 & 94.6 & 83.4 & 94.6 & 100.0 & 99.3 & 0.0 & 0.0 & 0.0 & 45.55 \\
Typed-token Transformer (ours) & 1 & 16 & 93.6 & 77.0 & 92.0 & 100.0 & 98.7 & 0.0 & 0.0 & 0.0 & 45.55 \\
Typed-token Transformer (ours) & 1 & 32 & \textbf{90.8} & \textbf{65.7} & \textbf{90.2} & 100.0 & \textbf{98.0} & 0.0 & \textbf{0.0} & 0.0 & 45.55 \\
Typed-token Transformer (ours) & 1 & 64 & 88.8 & 52.5 & 85.7 & 100.0 & 96.7 & 1.8 & 0.0 & 0.0 & 45.55 \\
Typed-token Transformer (ours) & 1 & 128 & 86.8 & 34.6 & \textbf{72.3} & 100.0 & 94.6 & 0.0 & 0.0 & 0.0 & 45.55 \\
\bottomrule
\end{tabular}
}%
\end{table*}

The typed representation separates positional continuity from aggregate field
accuracy. At $H=1$ it obtains 97.9\% semantic, 94.9\% active-field, and 99.1\%
head-position accuracy, while all dense predictors have at most 2.7\%
head-position accuracy. Its structural contradiction rate is zero, while every
dense predictor is contradictory on every evaluated tick. At $H=32$ it retains
90.8\% semantic and 90.2\% head-position accuracy with no collapse. The
full-state exact score reaches 41.7\% at $H=1$, with one-step mismatches
arising from simultaneous multi-food updates at $N=8$ rather than structural
contradictions.

Additional runs clarify how the model responds to data weighting, token order,
and training length. Extending the same-seed budget from 20{,}000 to 100{,}000
updates raises $H=1$ exactness to 45.8\% and head-position accuracy at $H=32$ and $H=128$ to 96.4\% and 83.0\%, respectively. Reordering player tokens before food raises $H=1$
exactness from 41.7\% to 45.8\%, with $H=32$ head-position accuracy changing
from 90.2\% to 80.4\%. Weighting transitions by observed spawn count leaves
$H=1$ exactness unchanged and reduces active accuracy. We retain the uniform
20{,}000-update checkpoint for the matched comparison and report the other
runs as fixed diagnostics rather than selecting metrics post hoc.

\setcounter{figure}{0}\setcounter{table}{0}
\section{Scaling and stability studies}
\label{app:scaling}
\subsection{Cross-game results}
\label{supp:three-game}

\begin{figure}[t]
\centering
\includegraphics[width=\columnwidth]{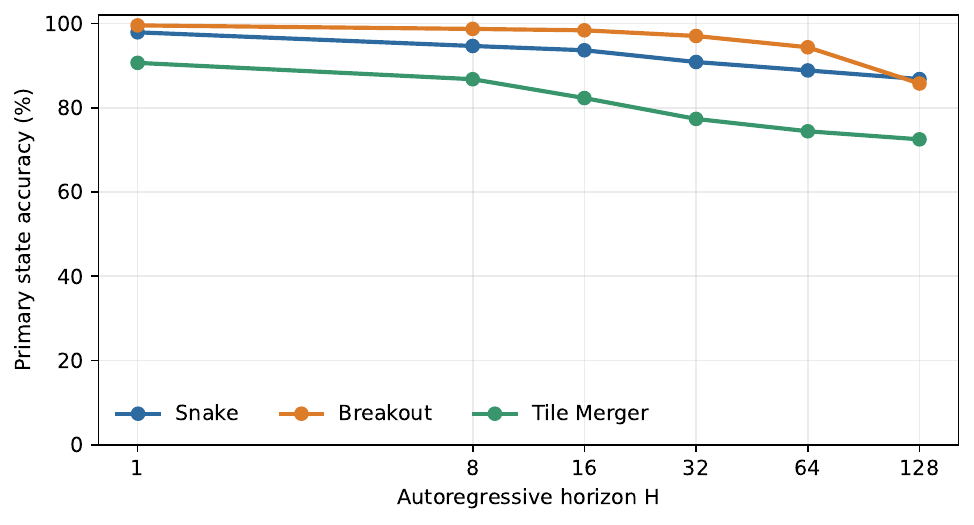}
\caption{Held-out autoregressive state accuracy across three games.}
\label{fig:typed-token-three-game}
\end{figure}

\begin{figure*}[t]
\centering
\includegraphics[width=0.98\textwidth]{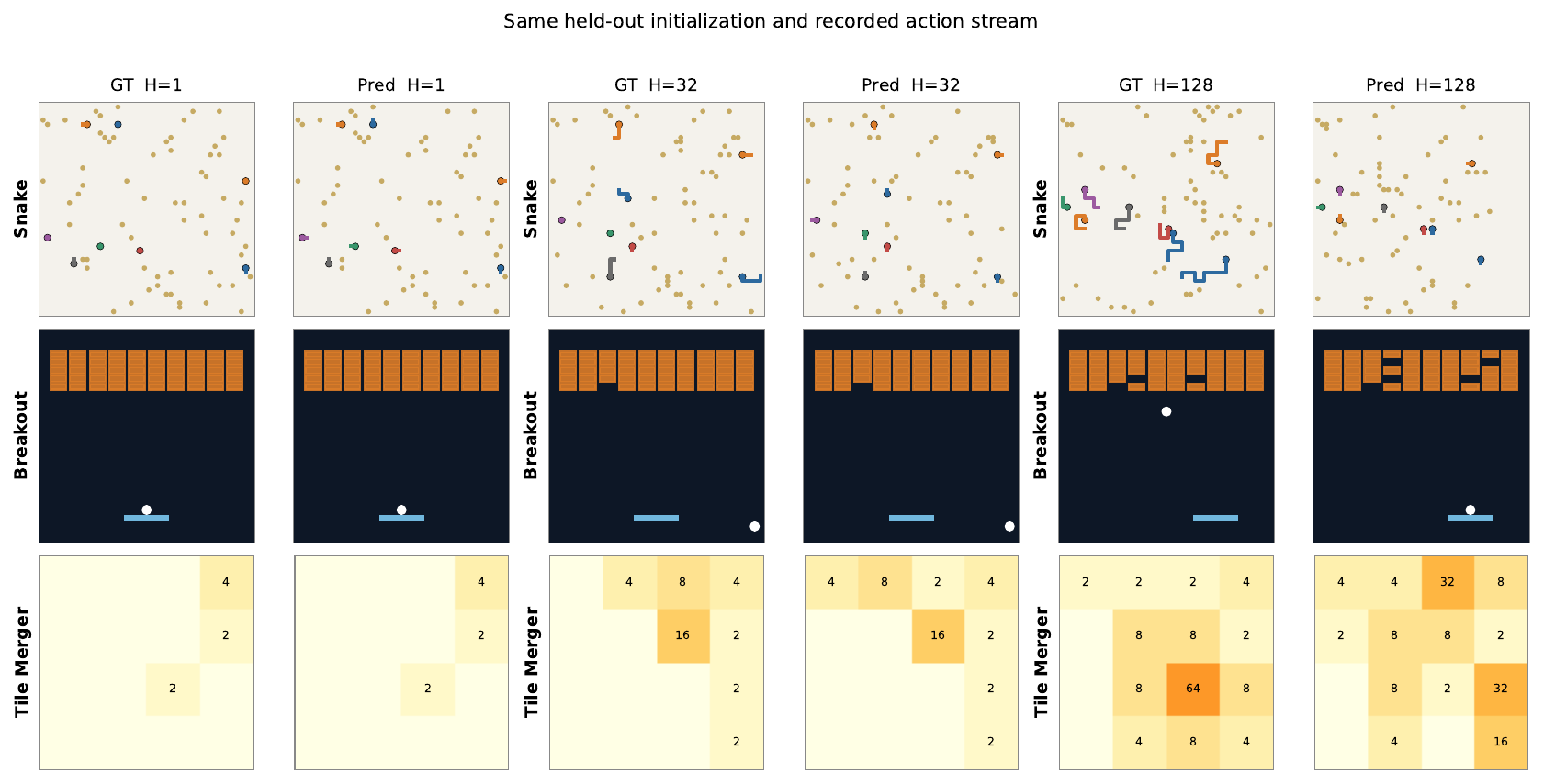}
\caption{Predicted and reference state sequences across three games.}
\label{fig:typed-token-three-game-rollouts}
\end{figure*}

Keeping width 256, six layers, eight heads, batch size 8, 20{,}000 updates,
and the seed fixed, we change only the declarative codec and game data.
Breakout contributes 3{,}466 training and 855 validation transitions. Tile
Merger contributes 5{,}729 and 1{,}916.  Every validation window starts from
its recorded state, feeds recorded actions and compatibility exogenous input,
then feeds predicted states back through $H=128$ (all 220 eligible Breakout
and 900 eligible Tile Merger windows). \Cref{tab:typed-token-three-game}
reports the endpoint metrics. \Cref{fig:typed-token-three-game} shows their
horizon trends, and \cref{fig:typed-token-three-game-rollouts} shows predicted
and reference states.

\begin{table}[t]
\centering
\small
\caption{Typed-token state prediction across three games.}
\label{tab:typed-token-three-game}
\begin{tabular*}{\columnwidth}{@{\extracolsep{\fill}}llrrrrr@{}}
\toprule
Game & Metric & $H$ & State acc. & Full exact & Invalid & Windows \\
\midrule
Snake & Field semantic & 1 & 97.9 & 41.7 & 0.0 & 24 \\
Snake & Field semantic & 32 & 90.8 & 0.0 & 0.0 & 24 \\
Snake & Field semantic & 128 & 86.8 & 0.0 & 0.0 & 24 \\
Breakout & Active token & 1 & 99.5 & 84.5 & 0.0 & 220 \\
Breakout & Active token & 32 & 97.0 & 43.2 & 0.0 & 220 \\
Breakout & Active token & 128 & 85.7 & 3.2 & 0.0 & 220 \\
Tile Merger & Active token & 1 & 90.6 & 0.0 & 0.0 & 900 \\
Tile Merger & Active token & 32 & 77.3 & 0.0 & 0.0 & 900 \\
Tile Merger & Active token & 128 & 72.5 & 0.0 & 0.0 & 900 \\
\bottomrule
\end{tabular*}
\end{table}

The state sequences use the first validation window for each game in manifest
order. Every prediction and reference pair uses the same initial state, action
stream, and horizon, independent of visual quality.

Breakout reaches 99.5\% active-token accuracy and 84.5\% full-state exactness
at $H=1$, and 85.7\% active-token accuracy with 3.2\% full-state exactness at
$H=128$. Every decode remains structurally valid. Tile Merger retains 72.5\%
active-token accuracy and structural validity at $H=128$, while its full-state
exact score is zero across the reported horizons. These results separate
structural validity from stochastic-grid exactness and show the same
interface supports closed-loop models for entity-table, ordered-entity, and
stochastic-grid states. The Breakout and Tile Merger datasets use
compatibility exogenous channels because their recorders precede engine-native
exogenous tokens. Breakout uses a projection from the next tick, and Tile
Merger uses the recorded spawn, making each logged transition deterministic.

\subsection{Rendering across games}
\label{supp:cross-game-results}
\label{supp:cross-game-rendering}

The cross-game study tests the shared state and camera interface across visual
domains. Population-scale Snake, Crate Pusher, and Pac-Man pair 1,024-entity
states with separately trained renderers. The interface stays fixed across
changes in game geometry, object semantics, and appearance.

The same state and camera contract is paired with separately trained 2D and
2.5D renderers. On held-out reference states, the two renderers obtain PSNR
33.54 and 30.12 with SSIM 0.9922 and 0.9855 for Snake. They obtain PSNR 35.38
and 34.39 with SSIM 0.9970 and 0.9898 for Crate Pusher. They obtain PSNR 34.88
and 32.72 with SSIM 0.9952 and 0.9795 for Pac-Man. These measurements isolate
view synthesis from recurrent state prediction.

All six renderers exceed SSIM 0.97 on reference states. The SRSC analysis
separates this renderer quality from state evolution. Across the three
population-scale games, the state and camera interface is shared, with
game-specific weights for each renderer.

\subsection{Long-horizon stability}
\label{supp:stability}

\begin{figure}[t]
\centering
\includegraphics[width=\columnwidth]{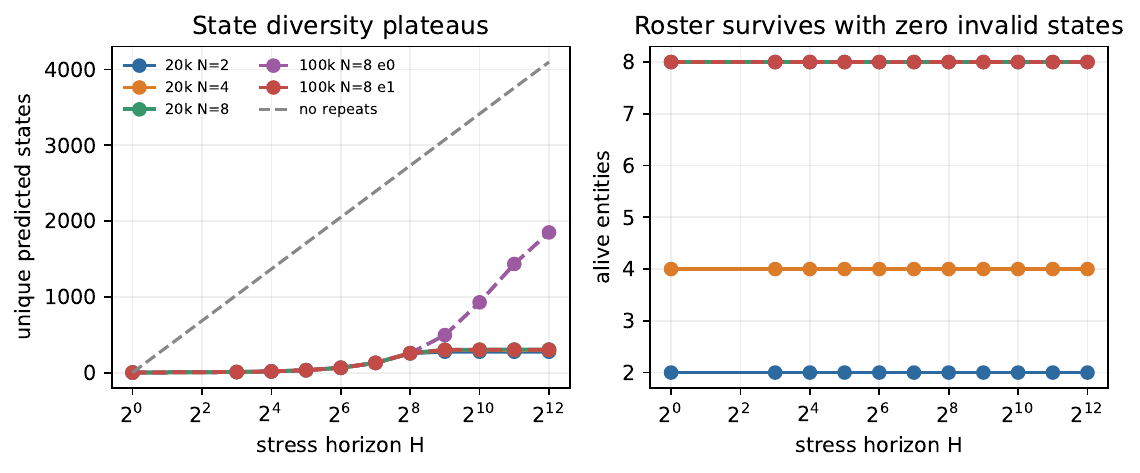}
\caption{Structural validity, roster survival, and state diversity through
$H=4096$ on Snake.}
\label{fig:snake-long-stability}
\end{figure}

Recorded Snake trajectories provide 159 reference transitions. We extend the
horizon with a separate stress test from held-out initial states, using a fixed
no-op action stream and explicit empty-spawn tokens. The test measures
structural decodability, roster survival, collapse, and recurrence independently
of a reference continuation. \Cref{tab:snake-long-stability} reports the
endpoint measurements, and \cref{fig:snake-long-stability} follows them over
the rollout.

\begin{table}[t]
\centering
\small
\caption{Long-horizon Snake results without ground-truth continuation.}
\label{tab:snake-long-stability}
\begin{tabular*}{\columnwidth}{@{\extracolsep{\fill}}lrrrrr@{}}
\toprule
Run & $H$ & Alive & Unique states & Invalid & Latency ms \\
\midrule
20k N=2 & 128 & 2 & 129 & 0.0 & 867.4 \\
20k N=2 & 4096 & 2 & 277 & 0.0 & 774.2 \\
20k N=4 & 128 & 4 & 129 & 0.0 & 834.2 \\
20k N=4 & 4096 & 4 & 301 & 0.0 & 740.7 \\
20k N=8 & 128 & 8 & 129 & 0.0 & 866.0 \\
20k N=8 & 4096 & 8 & 306 & 0.0 & 764.9 \\
\bottomrule
\end{tabular*}
\end{table}

Through $H=4096$, populations 2, 4, and 8 retain full rosters, all 64 food
slots, zero structurally invalid steps, and no collapse. The trajectories visit
between 277 and 306 unique predicted states before entering a recurrent
attractor, so state diversity complements the structural and collapse metrics.
With the 100{,}000-update checkpoint, two initializations produce 1{,}850 and
302 unique states. The longer training run can substantially extend diverse
evolution, while the difference between starts shows the effect of
initialization.

\subsection{Large-scale and asynchronous rollouts}
\label{supp:large-scale-async}

The Logic Engine advances one shared world, after which the Rendering Engine
serves the requested cameras from that prediction. We evaluate these stages
with 1,024 entities and during short gaps in authoritative updates.

\paragraph{One world transition and many view requests.}\label{supp:stage-costs}
The selected Snake state contains 1,024 player records, 4,096 food bucket
records, and one global record. The Logic Engine therefore predicts 5,121
records once per world tick. A controlled renderer sweep on one NVIDIA H100
then fixes the predicted state at tick 140 and simulates client-side rendering
for varying numbers of requested views over
$C_r\in\{1,4,16,64,256,1024\}$. The measured renderer cost
grows from a mean of 189.6\,ms for one requested view to 842.4\,ms for
1,024 requested views. The learned CNN component grows from 4.4\,ms to
379.7\,ms over the same sweep.

Increasing the number of requested views changes the rendering workload while
leaving the learned world transition unchanged. The cameras query one
prediction rather than maintaining separate recurrent worlds.

\paragraph{Client prediction under missing updates.}\label{supp:client-results}
We simulate an authoritative update stall in the 1,024-entity Snake rollout.
Windows begin at three held-out anchors and last from one to eight ticks. One
condition supplies the complete recorded joint-input stream. The second
supplies only the action of the evaluated player, assigns no-op actions to the
other players, and withholds future food spawns. Both conditions advance the
complete typed state from the last authoritative update.

With complete joint inputs, the local rollout matches the authoritative
in-view state in all 24 windows. With only the local action, the
selected avatar remains exact in all 24 windows while unseen actions and
events progressively alter the surrounding world. The shared state therefore
supports exact local continuation when the required world inputs are known and
keeps local responsiveness distinct from prediction of unobserved global
events.

\setcounter{figure}{0}\setcounter{table}{0}
\section{Netcode and diagnostics}
\label{app:netcode}
\subsection{Client prediction}
\label{supp:client-prediction}

When a server update is delayed, the client continues to render frames by
advancing its latest authoritative state locally. This section measures how
well that local prediction holds up before the next server state arrives.

\paragraph{Setup.}
The evaluation uses Snake at $\ntick=1024$ with the same $256\times6\times8$
typed-token Logic Engine used in the main experiments. For each held-out
episode, three anchor ticks ($t\in\{100,300,600\}$) serve as the last
received authoritative state. From each anchor, the client advances the full
typed state for up to $k=8$ ticks, then compares every predicted tick against
the recorded authoritative state at the same tick.

\paragraph{Variants.}
The \emph{oracle joint-input} variant supplies the local simulation with every player's
recorded action and the recorded exogenous spawns. It isolates the transition
accuracy when the complete input stream is available. The \emph{client}
variant supplies only the evaluated client's recorded actions, extrapolates
the other 1{,}023 players with no-op actions, and leaves spawn lists empty.
All other token fields, including the tick counter, retain their recorded
values.

\paragraph{Metrics.}
In-view agreement is the multiset intersection-over-union of visible
$(\text{kind},x,y)$ objects between the predicted and authoritative states,
both projected through the authoritative camera of the evaluated client.
Own-avatar displacement is the $L_1$ distance between the predicted and
authoritative head positions of the evaluated client. Each agreement value
at a given $k$ averages the three authoritative anchors.

\paragraph{Interpretation.}
The oracle joint-input variant holds agreement 1.000 through $k=8$ at every
anchor. Within a 400\,ms window the learned transition contributes no in-view
divergence, so the client variant's decay (0.815 at $k=1$ to 0.429 at $k=8$)
measures the effect of unknown remote inputs and withheld exogenous spawns,
which reconciliation corrects when a new server state arrives. Under the
evaluated anchors the local avatar incurs zero displacement, so no position
correction is needed for that player when the next server state arrives.

\paragraph{Execution regime.}
The client test advances the full typed state. At the matched two-player scale,
the transition takes 45.6\,ms and fits a 20\,Hz tick budget. At
$\ntick=1024$, the timing corresponds to a full-state step on one GPU.
System-level concerns such as neighborhood projection, transport policy, and
wall-clock optimization are not varied in this correctness measurement.

\subsection{SRSC protocol and results}
\label{supp:srsc}

Standard image metrics such as PSNR, SSIM, and LPIPS measure pixel-level or
perceptual similarity, but a frame that looks similar to the reference can
still be semantically wrong, with incorrect entity counts, positions, identities, or
interaction outcomes. Conversely, a rendered frame that differs in style or
sharpness may be semantically correct. State-Referenced Semantic Correctness
(SRSC) fills this gap by using a vision-language model (VLM) to judge whether
the semantic content of a rendered frame matches the reference, independent of
appearance.

\paragraph{Evaluation axes.}
A key advantage of \methodname{}'s architecture is that the explicit typed
state lets us evaluate the Logic Engine and Rendering Engine separately.
Video-based world models entangle dynamics and appearance in a shared visual
latent, and do not expose a comparable explicit interface for separating
errors in world transition from errors in view synthesis.  Because \methodname{} predicts a
typed state before rendering, we can replace either component with its
ground-truth counterpart and isolate the source of degradation.  This gives
four axes, each written as (reference, candidate).

\begin{itemize}
\item \textbf{Logic axis}:
  \\ reference = teacher render of the ground-truth state $s_t$,\\
  \\ candidate = teacher render of the predicted state $\hat{s}_t$.
  Both sides use the same teacher (reference) renderer, so any semantic
  difference is caused by the Logic Engine.
\item \textbf{Renderer-GT axis}:
  \\ reference = teacher render of $s_t$,\\
  \\ candidate = learned render of $s_t$.
  Both sides condition on the same ground-truth state, isolating the Rendering
  Engine's contribution without contamination from state error.
\item \textbf{Renderer-Pred axis}:
  \\ reference = teacher render of $\hat{s}_t$,\\
  \\ candidate = learned render of $\hat{s}_t$.
  Both sides condition on the same predicted state, isolating the Rendering
  Engine's contribution under the typical (degraded) input distribution.
\item \textbf{End-to-end axis}:
  \\ reference = teacher render of $s_t$,\\
  \\ candidate = learned render of $\hat{s}_t$.
  A full learned rollout frame is compared against the teacher reference.  This measures
the complete pipeline.
\end{itemize}

\paragraph{Protocol and prompt.}
The VLM judge is Qwen3.5-27B served with tensor parallelism. The reference
image is a symbolic semantic raster where each cell is colored by entity type (e.g., snake head, body, food, wall, empty), entity identity (player index), and team affiliation, with positions given by the raster cell coordinates. The candidate image is either a learned-renderer
RGB frame or a symbolic raster of the predicted state, depending on the axis.

The VLM receives the following prompt.

\begin{lstlisting}
You are a semantic game-state judge. Two images are provided:
REFERENCE (ground-truth symbolic raster) and CANDIDATE (learned
render or predicted-state raster).

1. For each image separately, inventory every visible entity by
   class (snake head, body, food, wall, empty, ...), identity
   (player index), team, and owner.
2. Assign each entity to a coarse spatial sector
   (left/center/right, top/middle/bottom) and note relative
   spatial relations.
3. Establish correspondence between the two inventories.
4. Check object states (alive/dead, moving/stationary),
   interactions (collision, collection), and list any
   hallucinated or missing entities.

Ignore differences in artistic style, texture, lighting, and
image sharpness. Focus only on semantic content: what entities
exist, where they are, and what state they are in.

Return a JSON object with exactly these fields:
  - "consistent": true or false
  - "score": integer 0-100
  - "mismatches": array of {"type": str, "description": str}
    (allowed types: missing_entity, hallucinated_entity,
     wrong_position, wrong_identity, wrong_team,
     wrong_object_state, wrong_interaction, wrong_geometry,
     uncertain_or_occluded)
  - "brief_reason": short auditable summary
\end{lstlisting}

The VLM returns a structured JSON response with four fields
\begin{itemize}
\item \texttt{consistent} (boolean): whether the semantic content agrees;
\item \texttt{score} (integer, 0--100): confidence-weighted semantic agreement;
\item \texttt{mismatches} (array): enumerated discrepancies drawn from a fixed
taxonomy (missing entity, hallucinated entity, wrong position, wrong identity,
wrong team, wrong object state, wrong interaction, wrong geometry, uncertain
or occluded)
\item \texttt{brief\_reason} (string): auditable summary of the inventory and
correspondence.
\end{itemize}

The table below reports the \textbf{mean SRSC score} (average of the
\texttt{score} field, range 0--100).  The binary version
(\texttt{SRSC\_binary}, fraction of frames where \texttt{consistent=true}) is
not used in this report because the finite sample sizes would produce
overly coarse granularity. The mean score provides a more informative
comparison.  The ``Pairs'' column in the table lists the
image-pair count and binary denominator for each axis.

Each evaluation uses deterministic greedy decoding (temperature~0) with a
pinned random seed (\texttt{20260722}).  The structured JSON response schema
enforces the exact field set, and every response is validated against that
schema before it enters the aggregate.  The prompt, schema, taxonomy,
visualizer specification, judge model snapshot, and service endpoint are
recorded and hashed before evaluation begins, and any change invalidates the
protocol binding.

\begin{table}[t]
\centering
\small
\caption{SRSC results for logic, rendering, and end-to-end predictions.
Pac-Man ends at $H=300$.}
\label{tab:srsc-axes}
\begin{tabular*}{\columnwidth}{@{\extracolsep{\fill}}llrrrrrrr@{}}
\toprule
Game & Axis & Scenes & $H{=}1$ & $H{=}100$ & $H{=}300$ & $H{=}600$ & $H{=}1200$ & Pairs \\
\midrule
Snake & Logic & 20 & 100.0 & 91.2 & 33.8 & 1.2 & 1.2 & 20 \\
 & Renderer-Pred & 20 & 53.8 & 76.9 & 60.6 & 48.8 & 30.6 & 40 \\
 & End-to-end & 20 & 55.6 & 51.2 & 31.2 & 7.5 & 2.5 & 40 \\
\addlinespace
Crate Pusher & Logic & 20 & 78.1 & 10.0 & 13.8 & 16.9 & 12.5 & 20 \\
 & Renderer-Pred & 20 & 78.8 & 7.5 & 0.0 & 12.5 & 5.0 & 40 \\
 & End-to-end & 20 & 69.4 & 5.0 & 0.0 & 5.0 & 2.5 & 40 \\
\addlinespace
Pac-Man & Logic & 6 & 100.0 & 60.0 & 10.0 & N/A & N/A & 6 \\
 & Renderer-Pred & 6 & 100.0 & 100.0 & 65.0 & N/A & N/A & 12 \\
 & End-to-end & 6 & 100.0 & 70.0 & 10.0 & N/A & N/A & 12 \\
\midrule
\multicolumn{9}{l}{\emph{Renderer-GT axis (pure renderer isolation, Logic Engine removed)}} \\
Crate Pusher & Renderer-GT & 10 & 100.0 & 75.0 & 100.0 & 100.0 & 100.0 & 10 \\
Snake & Renderer-GT & 10 & 68.8 & 71.2 & 63.8 & 75.0 & 66.2 & 10 \\
\bottomrule
\end{tabular*}
\end{table}

The ``Scenes'' column reports the number of independent starting states
(episode--timestep combinations) sampled for each axis.  ``Pairs'' is the
total number of (reference, candidate) image pairs presented to the VLM.
The Logic axis uses one pair per scene (teacher vs.\ teacher rendering of the
same world), while axes that involve rendering use two pairs per scene (two
camera viewpoints).  Percentages in the table
are the mean of the 0--100 \texttt{score} field.

\paragraph{Results.}
The SRSC mean scores are reported in \Cref{tab:srsc-axes}.
On Snake, the Logic axis
stays above 90\% through $H{=}100$ and degrades gradually at longer horizons.
The Renderer-Pred axis holds moderate scores across all horizons (30--77\%), while
the end-to-end scores follow the Logic axis decay. On Crate Pusher, all
axes drop faster, reflecting the higher interaction
density in that game. On Pac-Man, the Renderer-Pred axis remains
high (100\% at $H{=}1$ and 65\% at $H{=}300$), while the Logic and
end-to-end scores decay with horizon. The bottom rows report a control
condition (Renderer-GT axis), where the renderer conditioned on ground-truth
states---a setting that removes the Logic Engine entirely---achieves
near-perfect or high scores at all horizons without a clear monotonic decline.
This indicates that the Rendering Engine itself does not accumulate semantic
error with rollout horizon when the input state is held correct, and that the
decay observed in the full pipeline originates primarily in the learned
dynamics rather than in rendering.

The Renderer-GT results suggest that the Rendering Engine does not accumulate
substantial semantic drift over the evaluated horizons. The scores fluctuate
between 63.8 and 75.0 without a monotonic decline. Their offset from the
symbolic-raster ceiling reflects the representational gap between a categorical
semantic raster and a learned RGB rendering rather than temporal accumulation.
This separation of concerns is enabled by MASS's typed-state interface.

\end{document}